\documentclass[10pt,twocolumn,letterpaper]{article}

\usepackage[pagenumbers]{cvpr}

\newcommand{\red}[1]{{\color{red}#1}}

\usepackage{overpic}
\usepackage{pifont}
\usepackage[dvipsnames]{xcolor}
\usepackage{multirow}
\newcommand{\thRows}[1]{\multirow{3}*{#1}}

\newcommand{\tRows}[1]{\multirow{2}*{#1}}
\newcommand{\tCols}[1]{\multicolumn{2}{c}{#1}}
\newcommand{\figref}[1]{Fig.~\ref{#1}}
\newcommand{\tabref}[1]{Tab.~\ref{#1}}
\newcommand{\secref}[1]{Section~\ref{#1}}
\newcommand{\cmark}{\ding{51}}
\newcommand{\xmark}{\ding{55}}

\newcommand{\nameofdata}{AODRaw}
\newcommand{\myPara}[1]{\noindent\textbf{#1}}

\definecolor{cvprblue}{rgb}{0.21,0.49,0.74}
\usepackage[pagebackref,breaklinks,colorlinks,allcolors=cvprblue]{hyperref}

\title{Towards RAW Object Detection in Diverse Conditions}

\author{Zhong-Yu Li$^1$ \quad Xin Jin$^1$ \quad Boyuan Sun$^1$ \quad Chun-Le Guo$^1$ \quad Ming-Ming Cheng$^1$ \\
$^1$VCIP, School of Computer Science, Nankai University \\
\href{https://github.com/lzyhha/AODRaw}{\red{https://github.com/lzyhha/AODRaw}}
}

\begin{document}
\maketitle

\begin{abstract}
Existing object detection methods often consider sRGB input, which was compressed from RAW data using ISP originally designed for visualization. 
However, such compression might lose crucial information for detection, especially under complex light and weather conditions.
We introduce the \nameofdata{} dataset, which offers 7,785 high-resolution real RAW images with 135,601 annotated instances spanning 62 categories, 
capturing a broad range of indoor and outdoor scenes under 9 distinct light and weather conditions.
Based on \nameofdata{} that supports RAW and sRGB object detection, 
we provide a comprehensive benchmark for evaluating current detection methods.
We find that sRGB pre-training constrains the potential of RAW object detection 
due to the domain gap between sRGB and RAW, 
prompting us to directly pre-train on the RAW domain.  
However, 
it is harder for RAW pre-training to learn rich representations than sRGB pre-training 
due to the camera noise. 
To assist RAW pre-training, 
we distill the knowledge from an off-the-shelf model pre-trained on the sRGB domain. 
As a result, 
we achieve substantial improvements under diverse and adverse conditions 
without relying on extra pre-processing modules. 
\end{abstract}

\section{Introduction}
\label{sec:intro}

Real-world object detection is a fundamental task in computation vision. 
Significant advancements have been made in this field with public datasets like COCO~\cite{lin2015microsoft} and VOC~\cite{Everingham2009ThePV}. 
However, these datasets have predominantly focused on sRGB images, 
which lose some critical information compared to RAW images. 
The sensor first captures original RAW images with a high bit depth in a typical camera. 
An image signal processor (ISP) then compresses these RAW images into 8-bit sRGB images.
Unlike compressed sRGB images, 
RAW images retain a higher bit depth and thus preserve more distinguishable information, 
which is crucial for computer vision tasks, particularly in challenging light and weather conditions.
Moreover, inference on RAW images can reduce the time and power required by ISP.
Thus, RAW-based object detection has gained attention, especially in adverse conditions. 
However, exploration in this field remains limited. 

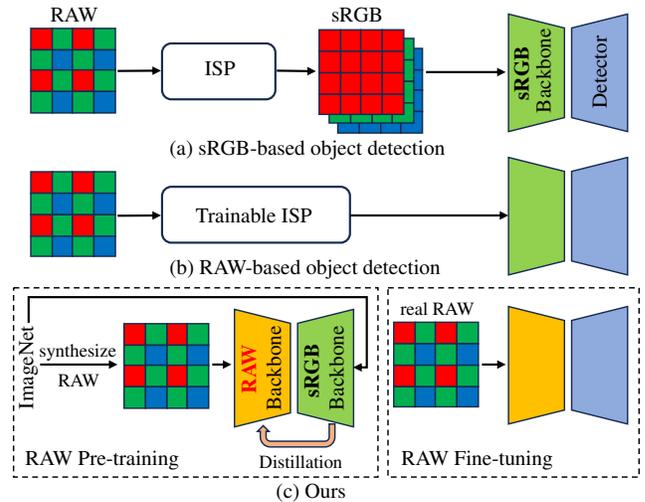
\begin{figure}
  \centering
  \vspace{10pt}
  \begin{overpic}[width=1.0\linewidth]{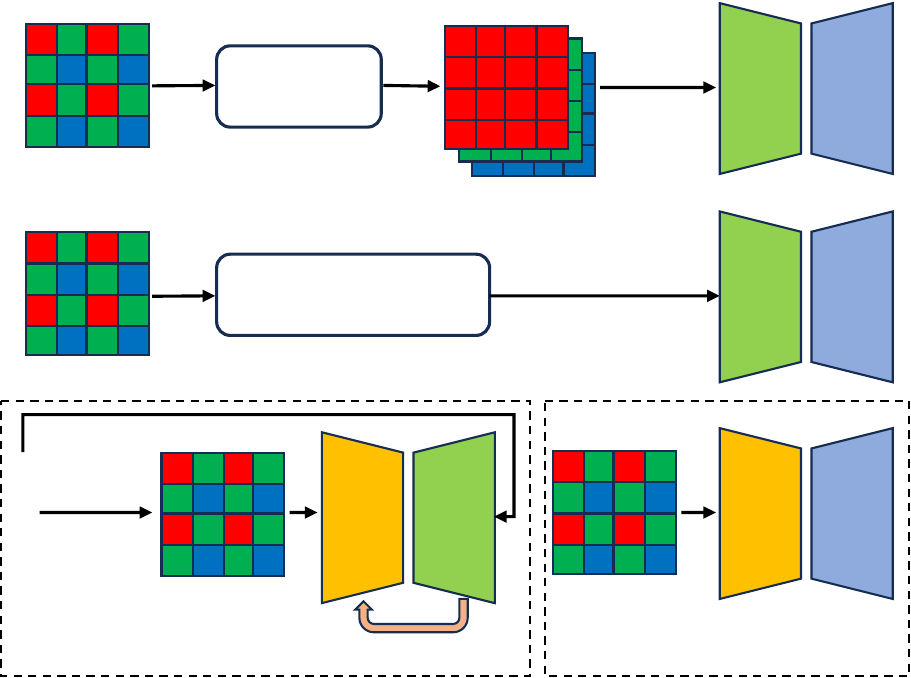}
    \put(6.0, 72.8){\footnotesize RAW}
    \put(50.8, 72.8){\footnotesize sRGB}
    \put(30.6, 64.2){\footnotesize ISP}
    \put(29.2, 40.8){\footnotesize Trainable ISP}
    \put(80.0, 60.3){\rotatebox{90}{\footnotesize {\textbf{sRGB}}}}
    \put(83.5, 58.3){\rotatebox{90}{\footnotesize Backbone}}
    \put(92.5, 59.0){\rotatebox{90}{\footnotesize Detector}}
    \put(25.0, 51.5){\footnotesize (a) sRGB-based object detection}
    \put(25.0, 32.5){\footnotesize (b) RAW-based object detection}
    \put(42.0, -3.2){\footnotesize (c) Ours}
    \put(2.0, 2.0){\footnotesize RAW Pre-training}
    \put(62.0, 2.0){\footnotesize RAW Fine-tuning}
    \put(4.2, 19.4){\scriptsize synthesize}
    \put(7.2, 14.6){\scriptsize RAW}
    \put(61.8, 26.0){\scriptsize real RAW}
    \put(1.0, 11){\rotatebox{90}{\footnotesize ImageNet}}
    \put(46.5, 12.8){\rotatebox{90}{\footnotesize {\textbf{sRGB}}}}
    \put(50.5, 10.8){\rotatebox{90}{\footnotesize Backbone}}
    \put(36.5, 13.5){\rotatebox{90}{\footnotesize {\textbf{\red{RAW}}}}}
    \put(40.5, 10.8){\rotatebox{90}{\footnotesize Backbone}}
    \put(39.3, 1.5){\scriptsize Distillation}
  \end{overpic}
  \caption{(a) Traditional sRGB-based object detection relies on 8-bit sRGB images, 
    which are compressed from RAW images and lose detailed information. 
    (b) Previous RAW-based methods utilize a trainable image signal processor~(ISP) 
    to adapt models pre-trained on the sRGB domain to the RAW domain. 
    (c) We pre-train models on the RAW domain, 
    achieving excellent performance on RAW object detection without requiring ISP modules.}
\label{fig:pipeline}
\end{figure}

The scarcity of relevant datasets is the key factor limiting the development of RAW-based object detection. 
However, collecting RAW images for object detection requires much more costs than sRGB images. 
For example, RAW images cannot be collected from picture websites like the sRGB-based dataset COCO~\cite{lin2015microsoft}. 
Thus, taking pictures requires a lot of labor, especially in rare weather conditions. 
Due to the limitations of collecting and annotating images, 
many RAW object detection methods~\cite{raw_adapter, Mosleh_2020_CVPR} 
rely on synthesizing RAW images 
that lack the authentic noise patterns and dynamic range. 
The real RAW datasets~\cite{omid2014pascalraw, Hong2021Crafting, CVPR2023_ROD} are also limited to the diversity. 
For example, LOD~\cite{Hong2021Crafting} and RAOD~\cite{CVPR2023_ROD} datasets are annotated with only 8 and 6 categories, respectively. 
Moreover, some datasets focus on outdoor scenes 
of daylight and low-light while neglecting other adverse conditions. 
Thus, existing methods have limited applications and cannot fully utilize the advantages of RAW images in handling adverse conditions.

To overcome these limitations, 
we propose a challenging dataset~({\textbf\nameofdata}) for \textbf{A}dverse condition 
\textbf{O}bject \textbf{D}etection 
with \textbf{RAW} images. 
\nameofdata~collect real RAW images from various indoor and outdoor scenes, 
where 2 light conditions, including daylight and low-light, 
and 3 weather conditions, including clear, rain, and fog, are considered. 
Because multiple light and weather conditions may co-occur, 
9 distinct conditions are collected. 
In total, we obtained 7,785 images and 135,601 annotated instances, with 6,504 images captured under adverse conditions. 
Meanwhile, our \nameofdata{} is annotated with 62 categories, significantly exceeding existing datasets. 
The diversity of scenes and semantics can further facilitate the development of RAW-based object detection in the real world. 
Furthermore, we evaluate existing RAW object detection methods~\cite{raw_adapter, CVPR2023_ROD} 
based on the \nameofdata{} dataset.

With \nameofdata, we aim to design a single model to detect objects across various conditions simultaneously, 
rather than training separate models for each condition in some previous approaches~\cite{raw_adapter}.
For RAW object detection, many methods usually transfer models pre-trained in the sRGB domain to the RAW domain using trainable adapters like neural ISP~\cite{raw_adapter}. 
However, the domain gap between sRGB and RAW impedes models from understanding the intricate information in RAW images, while the added adapters also introduce extra costs.
Some methods~\cite{CVPR2023_ROD} train models from scratch, yet limited data availability constrains performance and generalization. 

Differently, we explore pre-training on the RAW domain 
to reduce the domain gap between pre-training and fine-tuning, 
achieving notable improvements without any adapters. 
However, 
it is more difficult for models to learn high-quality representations 
from RAW images than sRGB images due to the camera noise. 
To this end, 
we propose to distill 
representations from an off-the-shelf model pre-trained on the sRGB domain to mitigate this difficulty. 
Taking ConvNext-T~\cite{liu2022convnet} and Cascade RCNN~\cite{Cai_2019}, 
sRGB-based object detection achieves 34.0\% AP on \nameofdata. 
RAW object detection improves the performance to 34.8\% AP 
through our RAW pre-training. 
To summarize, our main contributions are as follows:

\begin{itemize}
  \item We propose \nameofdata, 
  a high-quality dataset for RAW object detection under various light and weather conditions. 
  The dataset comprises diverse and complex images collected from various indoor and outdoor scenes.
  \item The \nameofdata{} supports research across multiple tasks, including RAW object detection and sRGB object detection under adverse conditions. 
  We evaluate the performance of existing object detection methods on these tasks.
  \item We pre-train models on RAW images via cross-domain distillation, 
  achieving significant improvements without needing adapters such as neural ISPs.
\end{itemize}

\section{Related Works}

\subsection{Object Detection}

Mainstream object detection methods can be divided into two categories, \ie, multi-stage and one-stage detectors. 
The multi-stage detectors, \eg, R-CNN series~\cite{Ren_2017,Sun_2021_CVPR,Cai_2019}, 
first generate region proposals and then refine them in subsequent stages. 
Cascade R-CNN~\cite{Cai_2019} further extends this process via multiple refinement stages, progressively improving localization and classification accuracy. 
Although these methods achieve high accuracy, they have the drawback of slow inference. 
One-stage methods, such as YOLO~\cite{yolox2021} and RetinaNet~\cite{lin2017focal}, 
directly predict object locations and categories, enabling faster inference but at a trade-off in precision. 
In addition, transformer-based approaches like Deformable DETR \cite{zhu2021deformable} have emerged and leverage self-attention mechanisms to model spatial relationships in the image while requiring longer training times. 
Because these methods are primarily designed for sRGB images, 
we further evaluates these classic methods in the RAW domain.

\begin{figure*}[t]
  \centering
  \includegraphics[width=\linewidth]{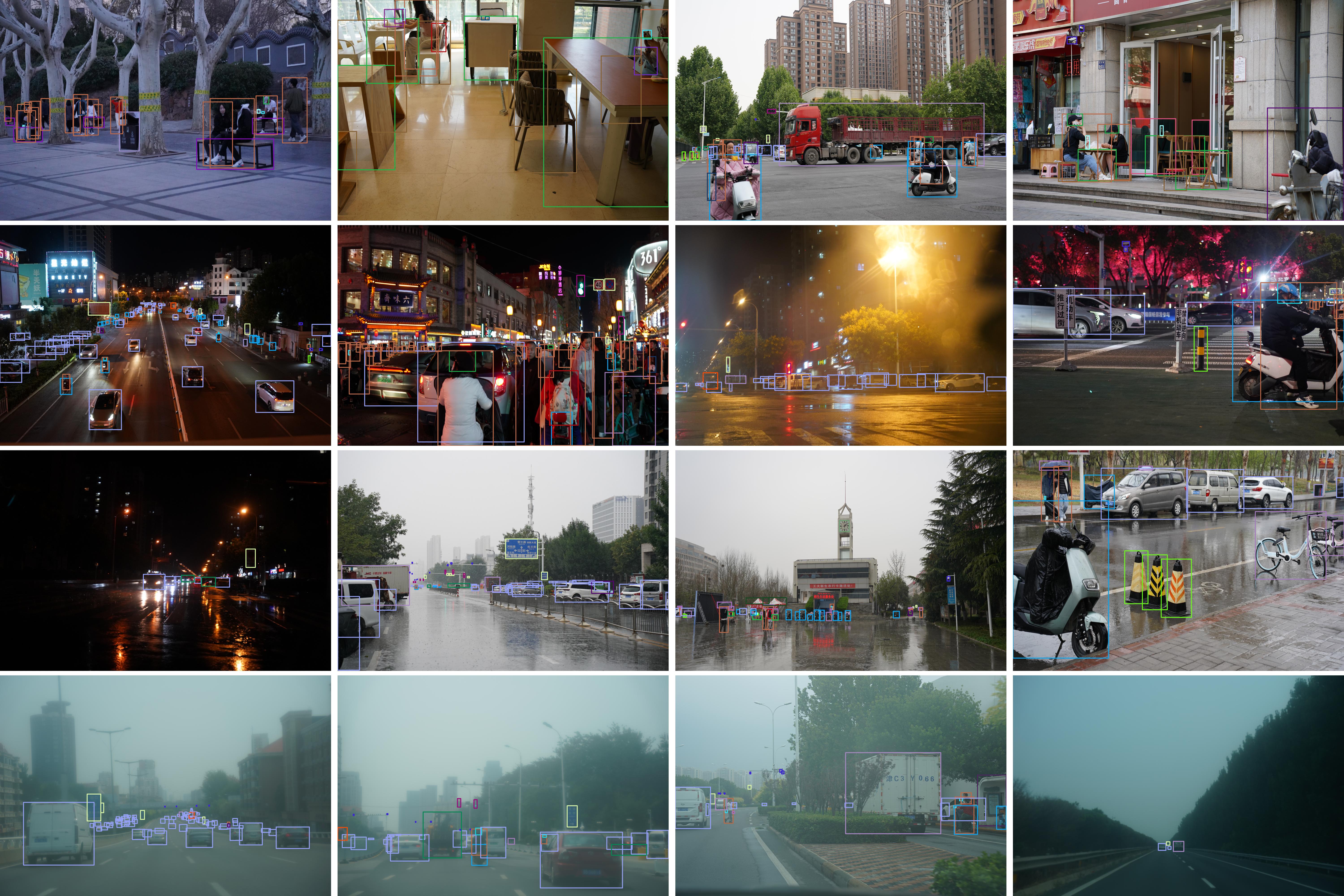}
  \caption{Example of the images in the \nameofdata. 
  From top to bottom, we show daylight, low-light, rain, and fog conditions, respectively. 
  A part of the images are taken under multiple conditions. 
  For example, the first one in the third row is taken in low-light and rain conditions. 
  More examples for each condition can be found in the supplementary material.
  }\label{fig:dataset_example}
\end{figure*}

\subsection{RAW Object Detection}

RAW object detection, which leverages unprocessed sensor data, 
has gained attention due to its potential in challenging light and weather conditions. 
However, the field lacks large-scale datasets for pre-training models on RAW images, which is essential for modern object detection methods. 
Thus, some methods~\cite{CVPR2023_ROD} train detectors from scratch using real-time detectors~\cite{yolox2021}. 
Due to camera noise and the limited quantity of RAW images, these methods may converge slowly and face limitations in performance. 
The other methods~\cite{8803607, Yoshimura_2023_ICCV, Yu_2021_ICCV} adapt models pre-trained on sRGB images to the RAW domain. 
Among them, some propose a differentiable image signal processor (ISP) \cite{wang2024adaptiveisp, Mosleh_2020_CVPR, raw_adapter, Dirtypixel, Guo_2024_CVPR, qin2022attention} for pre-processing RAW images. 
Fine-tuning models by synthesizing RAW images from COCO~\cite{2023lis,10415533} also helps mitigate the domain gap. 
However, the sRGB pre-training still limits the ability to understand RAW images that contain much more information than sRGB images, and the ISP modules add extra computation costs. 
Thus, we explore pre-training models on RAW images.

\subsection{RAW Object Detection Datasets}

Existing datasets for object detection, like COCO~\cite{lin2015microsoft}, primarily collect sRGB images. 
Although these datasets have driven significant progress in object detection, sRGB images lack the detailed information available in RAW images, which can be particularly beneficial in adverse conditions. 
Due to the scarcity of RAW datasets, many methods~\cite{raw_adapter, Mosleh_2020_CVPR} rely on synthetic datasets for RAW detection research.
Then, a few RAW datasets have been proposed. 
For example, \cite{zhang2023darkvision} collects images in low-light conditions, and \cite{Bijelic_2020_CVPR} collects driving images of high dynamic range in fog conditions. 
The PASCALRAW~\cite{omid2014pascalraw} dataset is collected similarly to the PASCAL VOC~\cite{Everingham2009ThePV}, providing 4,259 images in daylight. 
The LOD~\cite{Hong2021Crafting} captures 2,230 paired images in daylight and low-light conditions. 
RAOD~\cite{CVPR2023_ROD} has 25,207 annotated driving images but only covers 6 categories and 2 light conditions. 
Specific conditions, driving-focused scenes, or a narrow range of annotated categories limit these datasets. 

\begin{table*}
  \centering
  \setlength{\tabcolsep}{1.8mm}
  \begin{tabular}{lcccccl}
    \toprule
    Dataset & Resolution & Images & Categories & Instances & Instances per image & Conditions \\
    \midrule
    OnePlus~\cite{Yu_2021_ICCV} & $4640 \times 3480$ & 141 & 5 & 1,228 & 8.7 & 1 (low-light) \\
    PASCALRAW~\cite{omid2014pascalraw} & $600 \times 400$ & 4,259 & 3 & 6,550 & 1.5 & 1 (daylight) \\
    LOD~\cite{Hong2021Crafting} & $1200\times 800$ & 2,230 & 8 & 9,726 & 4.4 & 2 (daylight and low-light) \\
    RAOD~\cite{CVPR2023_ROD} & $2880 \times 1856$ & 25,207 & 6 & 237,379 & 9.4 & 2 (daylight and low-light) \\
    \midrule
    \nameofdata{} (Ours) & $6000 \times 4000$ & 7,785 & 62 & 135,601 & 17.4 & 9~(in \tabref{tab:image_per_scenario}) \\ \bottomrule
  \end{tabular}
  \caption{Comparison with existing RAW datasets.
   }
  \label{tab:existing_datasets}
\end{table*}

\begin{table*}
  \centering
  \setlength{\tabcolsep}{2.6mm}
  \begin{tabular}{lcccccccccc}
    \toprule
    & \tCols{Indoor} & \multicolumn{7}{c}{Outdoor} & \thRows{Total} \\
    \cmidrule(lr){2-3} \cmidrule(lr){4-10}
    Brightness & daylight & low-light & \multicolumn{4}{c}{daylight} & \multicolumn{3}{c}{low-light} \\
    \cmidrule(lr){2-2} \cmidrule(lr){3-3} \cmidrule(lr){4-7} \cmidrule(lr){8-10} 
    Weather & - & - & clear & fog & rain & fog+rain & clear & fog & rain \\
    Images & 477 & 1,210 & 804 & 1,110 & 1,252 & 244 & 1,842 & 325 & 521 & 7,785\\
    Instances & 4,992 & 10,195 & 18,575 & 23,636 & 24,107 & 5,381 & 37,282 & 4,513 & 6,920 & 135,601 \\
    \bottomrule
  \end{tabular}
  \caption{The number of images per condition.}
  \label{tab:image_per_scenario}
\end{table*}

\begin{figure*}[t]
  \centering
  \begin{subfigure}[t]{0.31\textwidth}
    \begin{overpic}[width=\linewidth]{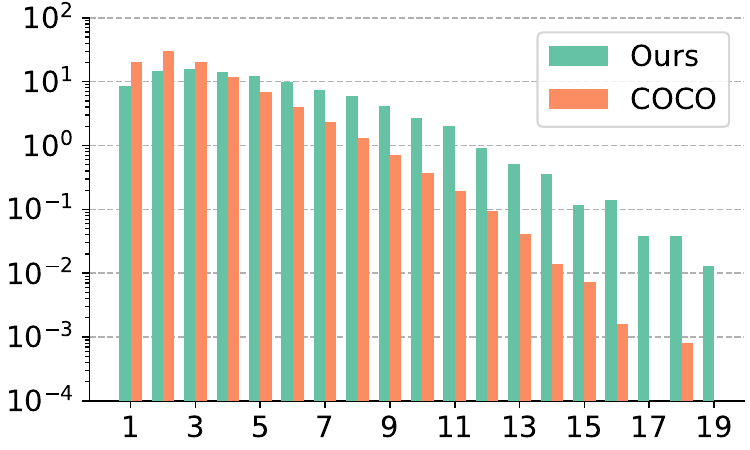}
      \put(36, -2){\scriptsize Number of categories}
      \put(-4, 9.5){\rotatebox{90}{\footnotesize Percent of images (\%)}}
    \end{overpic}
    \caption{Distribution of the number of categories in images.}
    \label{fig:category_per_image}
  \end{subfigure}
  \hfill
  \begin{subfigure}[t]{0.31\linewidth}
    \begin{overpic}[width=\linewidth]{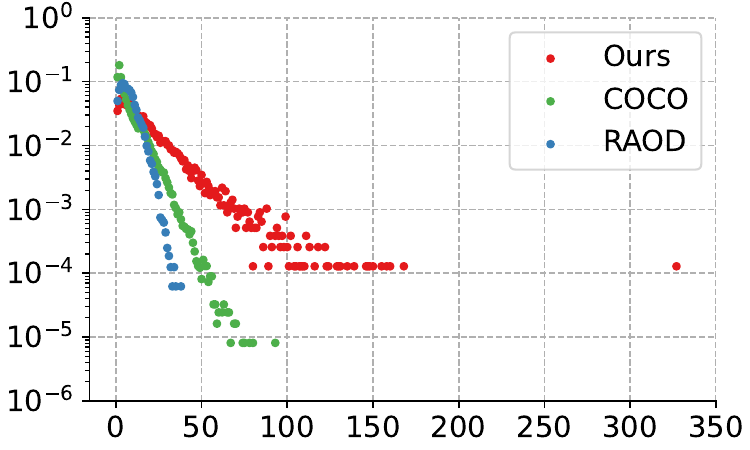}
      \put(37, -2){\scriptsize Number of instances}
      \put(-4, 9.5){\rotatebox{90}{\footnotesize Percent of images (\%)}}
    \end{overpic}
    \caption{Distribution of the number of instances in images.}
    \label{fig:instance_per_image}
  \end{subfigure}
  \hfill
  \begin{subfigure}[t]{0.31\linewidth}
    \centering
    \begin{overpic}[width=\linewidth]{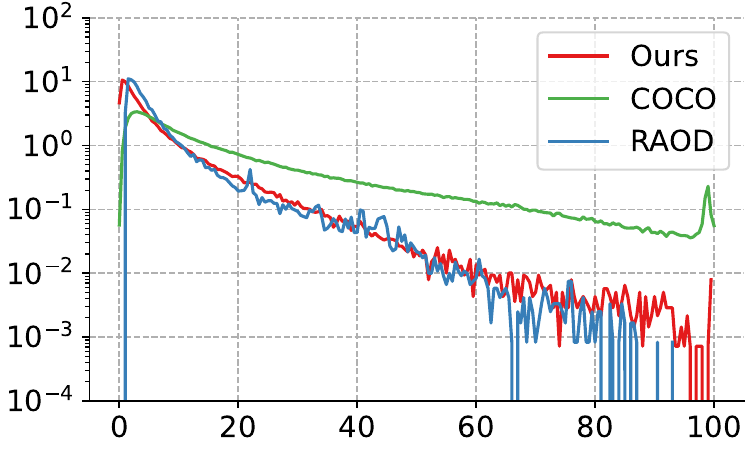}
      \put(40, -2){\scriptsize Relative box size}
      \put(-4, 6.5){\rotatebox{90}{\footnotesize Percent of instances (\%)}}
    \end{overpic}
    \caption{Relative bounding box size  $\sqrt{\frac{\rm box\ area}{\rm image\ area}}$.}
    \label{fig:instance_size}
  \end{subfigure}
  \begin{subfigure}[t]{0.31\linewidth}
    \centering
    \begin{overpic}[width=\linewidth]{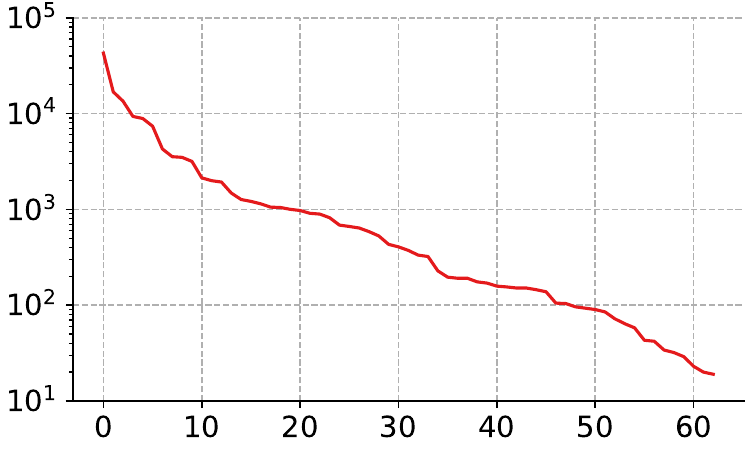}
      \put(33, -2){\scriptsize Sorted category index}
      \put(-4, 10){\rotatebox{90}{\footnotesize Number of instances}}
    \end{overpic}
    \caption{The number of instances per category.}
    \label{fig:instance_per_category}
  \end{subfigure}
  \hfill
  \begin{subfigure}[t]{0.31\linewidth}
    \centering
    \begin{overpic}[width=\linewidth]{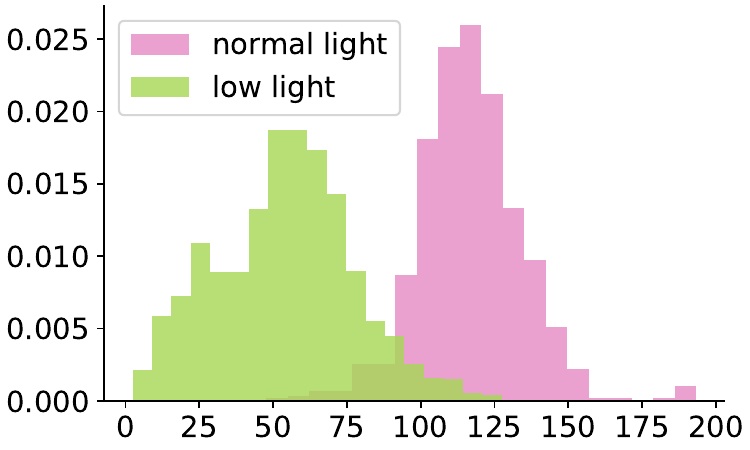}
      \put(46, -2){\scriptsize Brightness}
      \put(-4, 11){\rotatebox{90}{\footnotesize Probability density}}
    \end{overpic}
    \caption{Brightness distribution for outdoor scenes.}
    \label{fig:light_level}
  \end{subfigure}
  \hfill
  \begin{subfigure}[t]{0.31\linewidth}
    \centering
    \begin{overpic}[width=\linewidth]{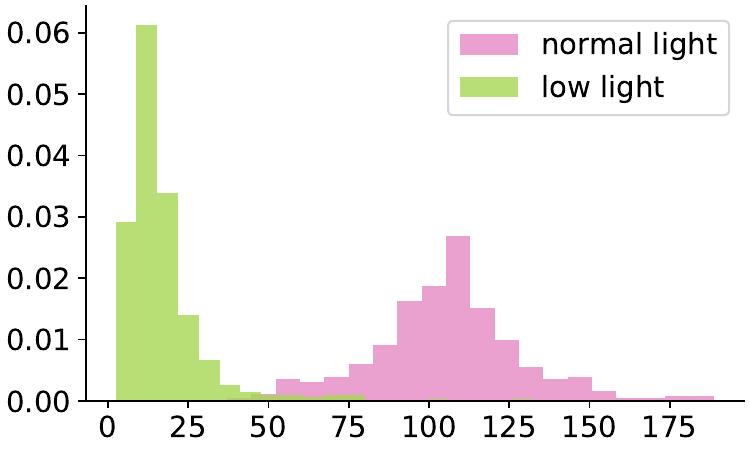}
      \put(46, -2){\scriptsize Brightness}
      \put(-4, 11){\rotatebox{90}{\footnotesize Probability density}}
    \end{overpic}
    \caption{Brightness distribution for indoor scenes.}
    \label{fig:light_level_indoor}
  \end{subfigure}
  \caption{Statistics indicate that our \nameofdata{} dataset contains increased category and instance diversity.}
  \vspace{-5pt}
\end{figure*}

\section{\nameofdata{} Dataset}
\label{sec:wod_dataset}

\subsection{Data Collection}

\myPara{Diverse conditions.} 
We construct a challenging dataset for RAW-based object detection under adverse and diverse conditions. 
Specifically, 
we consider 2 light conditions, \ie, daylight and low-light, 
and 3 weather conditions, \ie, clear, rain, and fog. 
For different light conditions, 
we capture both indoor and outdoor images. 
Because multiple conditions may co-occur, 
we finally collect 7,785 real RAW images and 
the corresponding sRGB images 
across 9 combined conditions, as shown in \tabref{tab:image_per_scenario}. 
For example, the image in the third row and first column of \figref{fig:dataset_example} is in rain and low-light conditions. 

\myPara{Data diversity.} 
In addition to condition diversity, we capture images across various cities and scenes to ensure broad data diversity. 
As shown in \figref{fig:dataset_example}, some images are taken in traffic scenes, while others cover gardens, universities, libraries, streets, and other indoor scenes.

\myPara{Data annotation.} 
We follow the annotation format of the COCO dataset~\cite{lin2015microsoft} to annotate bounding boxes in images across 62 categories commonly seen in daily life. 
\subsection{Data Analysis}

In this section, we analyze the \nameofdata{} dataset and compare it with two previous object detection datasets, 
\ie, COCO~\cite{lin2015microsoft} of sRGB object detection and RAOD~\cite{CVPR2023_ROD} of RAW object detection. 
In the supplementary material, 
we show the detailed analysis about each condition. 

\myPara{Diverse scenes.}
As summarized in \tabref{tab:image_per_scenario}, \nameofdata{} covers 9 conditions, 
including 2 light conditions, 3 weather conditions, and different combinations. 
Compared to existing datasets for RAW object detection, 
which mainly focus on outdoor scenes in daylight or low-light, as shown in \tabref{tab:existing_datasets}, 
\nameofdata{} has a greater diversity, presenting a more challenging task. 
Moreover, the scenes in \nameofdata{} are varied and complex. 
As shown in \figref{fig:category_per_image} and \figref{fig:instance_per_image}, 
images in \nameofdata~contain varying categories and instances, with up to 19 categories and 327 instances in an image. 
On average, 
there are 17.4 instances per image as shown in \tabref{tab:existing_datasets}, 
exceeding existing datasets.

\myPara{Increased category diversity.}
\nameofdata{} includes 62 categories, a significantly higher number than most existing RAW object detection datasets, as shown in \tabref{tab:existing_datasets}. 
Meanwhile, as shown in \figref{fig:instance_per_category}, the distribution of categories exhibits a long-tail pattern, further increasing the challenge of RAW object detection in this dataset. 

\myPara{Object scales.}
The instances in the \nameofdata{} vary widely in size, as shown in \figref{fig:instance_size}, with a notably larger proportion of small objects than in previous datasets. 
This variance requires the detectors to extract multi-scale representations, 
increasing the complexity of the detection task.

\begin{figure}[t]
  \centering
  \begin{subfigure}[t]{0.22\textwidth}
    \centering
    \begin{overpic}[width=\textwidth]{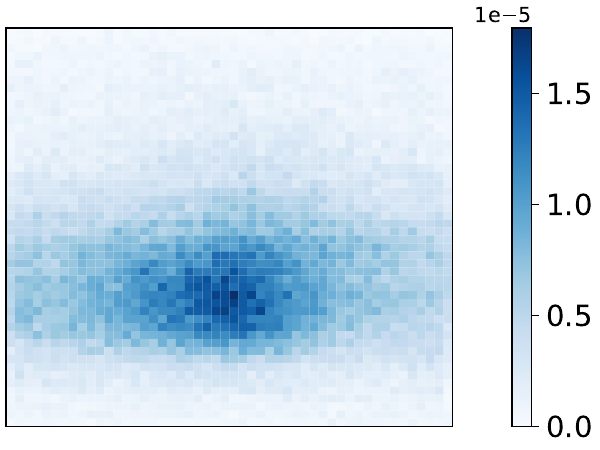}
    \end{overpic}
    \caption{Our \nameofdata.}
  \end{subfigure}
  \hspace{5pt}
  \begin{subfigure}[t]{0.22\textwidth}
    \centering
    \begin{overpic}[width=\textwidth]{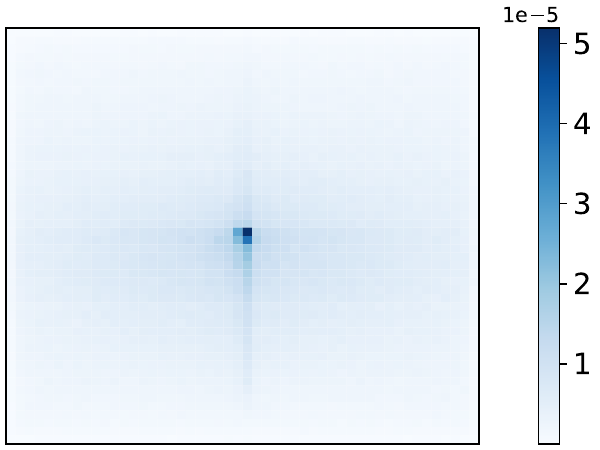}
    \end{overpic}
    \caption{COCO.}
  \end{subfigure}
  \caption{The distribution of object centers.}
  \label{fig:object_center}
\end{figure}

\myPara{Spatial distribution.}
The instances in the \nameofdata{} dataset are more uniformly distributed in the images, as shown in \figref{fig:object_center}. 
This uniform spatial distribution helps reduce spatial bias.
Additionally, there is a slight bias towards the bottom of images, as most images are captured in outdoor scenes.

\myPara{Light distribution.}
The distribution of lightness, calculated as the average gray value of sRGB images, is shown in \figref{fig:light_level} and \figref{fig:light_level_indoor}. 
The distribution reveals a broad range of light conditions in \nameofdata.

\subsection{Data Split}
The dataset is split for training and testing sets with a 7:3 ratio. 
To ensure that each split contains sufficient images of each condition, we split each condition individually and then merge the splitting results. 
As a result, we obtained 5,445 training images and 2,340 testing images, which contain 94,949 and 40,652 instances, respectively. 

\section{Benchmark}
\label{sec:benchmark}

\subsection{Implementation Details}

\myPara{Model training.}
We implement all object detection methods using a popular code base, mmdetection~\cite{mmdetection}. 
The models are trained for 48 epochs with a batch size of 16, except for Deformable DETR~\cite{zhu2021deformable}, 
which is trained for 100 epochs. 
Please refer to the supplementary material for more hyper-parameters. 
In addition, 
the RAW images are originally saved in the bayer pattern with the shape of $1\times H \times W$, where $H$ and $W$ mean the height and width of images. 
To be compatible with existing models, 
we transform the RAW images into $3\times H \times W$ using demosaicing 
following~\cite{CVPR2023_ROD}. 
Following \cite{10415533}, 
the RAW images are further processed through gamma correction for faster convergence.

\myPara{Image resolution.}
The images in the \nameofdata{} dataset are recorded at a resolution of $6000\times 4000$. 
It is unrealistic to feed such huge images into the detectors. 
Thus, we adopt two experiment settings: 1) 
down-sampling the images into a lower resolution of $2000\times1333$ 
following the approach in \cite{CVPR2023_ROD}, 
and 2) slicing the images into a collection of $1280\times 1280$ patches with a patch overlap of 300 and 
ignoring the objects whose IoU with the sliced images is lower than 0.4, resulting in 71,782 images and 417,781 instances.
The first setting supports faster training, but too tiny objects with an area of less than $32^2$ are ignored because they will disappear after down-sampling. 
The second requires more time for training, but it can fully use high-quality annotations and support tiny object detection~\cite{SODA}. 
In the following, we adopt down-sampling by default. 

\myPara{Evaluation protocol.}
We evaluate the models using the popular metric Average Precision~(AP)~\cite{mmdetection,lin2015microsoft}, along with AP$_{\rm 75}$ and AP$_{\rm 50}$ at the IoU threshold of 0.75 and 0.50. 
About AP$_{\rm s}$, AP$_{\rm m}$, and AP$_{\rm l}$ for small, medium, and large objects, we set object area ranges as [$0$, $128^2$), [$128^2$, $320^2$), and [$320^2$, $+\infty$), respectively, when using the setting of down-sampling images. 
When slicing images, the ranges are set as [$0$, $64^2$), [$64^2$, $160^2$), and [$160^2$, $+\infty$), respectively. 
To facilitate object detection in adverse conditions, 
we also report AP$_{\rm low}$, AP$_{\rm rain}$, and AP$_{\rm fog}$ for low-light, rain, and fog conditions, 
apart from AP$_{\rm normal}$ for the normal condition~(the combination of daylight and clear weather).

\begin{table*}
  \centering
  \setlength{\tabcolsep}{0.2mm}
  \begin{tabular}{llccccc|ccc|cccc}
    \toprule
    Method & Backbone & Pre-Train & Fine-Tune & AP & AP$_{\rm 50}$ & AP$_{\rm 75}$ & AP$_{\rm s}$ & AP$_{\rm m}$ & AP$_{\rm l}$ & AP$_{\rm normal}$ & AP$_{\rm low}$ & AP$_{\rm rain}$ & AP$_{\rm fog}$ \\
    \midrule
    Faster RCNN~\cite{Ren_2017} & ResNet-50~\cite{he2016deep} & \multirow{6}{*}{sRGB} & \multirow{6}{*}{sRGB} & 23.3 & 41.3 & 23.7 & 13.1 & 30.8 & 36.4 & 26.0 & 22.0 & 24.4 & 19.6 \\
    Retinanet~\cite{lin2017focal} & ResNet-50~\cite{he2016deep} & & & 19.1 & 33.6 & 19.2 & 10.1 & 26.6 & 29.5 & 21.5 & 17.8 & 19.2 & 16.5 \\
    GFocal~\cite{li2020generalized} & ResNet-50~\cite{he2016deep} & & & 24.2 & 40.3 & 24.7 & 13.3 & 31.9 & 37.0 & 26.5  & 22.3 & 24.1 & 21.1 \\
    Sparse RCNN~\cite{Sun_2021_CVPR} & ResNet-50~\cite{he2016deep} & & & 15.6 & 28.3 & 15.0 & 7.2 & 22.1 & 28.9 & 17.9 & 15.0 & 14.6 & 12.6 \\
    Deformable DETR~\cite{zhu2021deformable} & ResNet-50~\cite{he2016deep} & & & 16.6 & 31.9 & 15.6 & 7.7 & 23.9 & 30.1 & 18.3 & 15.2 & 16.4 & 13.1 \\
    Cascade RCNN~\cite{Cai_2019}& ResNet-50~\cite{he2016deep} & & & 25.6 & 41.4 & 26.4 & 13.7 & 32.4 & 38.3 & 27.3 & 23.8 & 24.7 & 20.4 \\
    \midrule
    Faster RCNN~\cite{Ren_2017} & Swin-T~\cite{liu2021Swin} & \multirow{6}{*}{sRGB} & \multirow{6}{*}{sRGB} & 28.4 & 50.1 & 28.8 & 15.6 & 35.9 & 42.6 & 32.0 & 26.0 & 27.2 & 23.3 \\
    Faster RCNN~\cite{Ren_2017} & ConvNeXt-T~\cite{liu2022convnet} & & & 29.7 & 51.7 & 30.1 & 17.1 & 37.3 & 45.4 & 33.1 & 28.3 & 27.1 & 24.4 \\
    GFocal~\cite{li2020generalized} & Swin-T~\cite{liu2021Swin} & & & 30.1 & 48.9 & 30.6 & 16.3 & 38.0 & 44.4 & 32.7 & 28.1 & 28.2 & 24.5 \\
    GFocal~\cite{li2020generalized} & ConvNeXt-T~\cite{liu2022convnet} & & & 32.1 & 49.9 & 33.6 & 18.7 & 39.9 & 49.5 & 35.2 & 30.3 & 31.8 & 26.0 \\
    Cascade RCNN~\cite{Cai_2019} & Swin-T~\cite{liu2021Swin} & & & 32.0 & 50.2 & 34.0 & 17.5 & 40.1 & 46.3 & 35.4 & 30.0 & 28.2 & 25.0 \\
    Cascade RCNN~\cite{Cai_2019} & ConvNeXt-T~\cite{liu2022convnet} & & & 34.0 & 52.7 & 36.3 & 19.3 & 40.8 & 52.1 & 37.0 & 31.5 & 32.9 & 27.2\\
    \midrule
    Faster RCNN~\cite{Ren_2017} & Swin-T~\cite{liu2021Swin} & \multirow{6}{*}{sRGB} & \multirow{6}{*}{RAW} & 28.1 & 50.0 & 28.2 & 16.0 & 35.7 & 42.6 & 30.7 & 26.5 & 26.2 & 22.0 \\
    Faster RCNN~\cite{Ren_2017} & ConvNeXt-T~\cite{liu2022convnet} & & & 29.4 & 51.3 & 29.6 & 16.3 & 37.6 & 44.4 & 32.7 & 27.3 & 29.2 & 24.6 \\
    GFocal~\cite{li2020generalized} & Swin-T~\cite{liu2021Swin} & & & 29.9 & 48.2 & 30.6 & 16.3 & 38.3 & 45.0 & 33.1 & 27.6 & 29.0 & 23.8 \\
    GFocal~\cite{li2020generalized} & ConvNeXt-T~\cite{liu2022convnet} & & & 31.5 & 50.0 & 32.9 & 17.9 & 39.5 & 48.4 & 34.9 & 29.4 & 32.2 & 26.7 \\
    Cascade RCNN~\cite{Cai_2019} & Swin-T~\cite{liu2021Swin} & & & 31.7 & 49.8 & 32.8 & 17.7 & 39.7 & 47.8 & 35.3 & 29.8 & 28.6 & 23.9 \\
    Cascade RCNN~\cite{Cai_2019} & ConvNeXt-T~\cite{liu2022convnet} & & & 33.7 & 52.0 & 35.9 & 18.6 & 41.7 & 51.3 & 36.8 & 31.3 & 31.3 & 27.2 \\
    \midrule
    Faster RCNN~\cite{Ren_2017} & Swin-T~\cite{liu2021Swin} & \multirow{6}{*}{RAW} & \multirow{6}{*}{RAW} & 28.6 & 50.2 & 28.5 & 15.6 & 36.9 & 43.1 & 32.1 & 26.7 & 27.6 & 23.2 \\
    Faster RCNN~\cite{Ren_2017} & ConvNeXt-T~\cite{liu2022convnet} & & & 30.2 & 52.3 & 31.0 & 17.0 & 39.1 & 46.9 & 33.8 & 27.7 & 30.2 & 26.6  \\
    GFocal~\cite{li2020generalized} & Swin-T~\cite{liu2021Swin} & & & 30.7 & 49.7 & 31.8 & 17.2 & 39.4 & 47.4 & 33.7 & 28.6 & 28.5 & 25.3 \\
    GFocal~\cite{li2020generalized} & ConvNeXt-T~\cite{liu2022convnet} & & & 32.1 & 50.4 & 33.4 & 17.7 & 40.6 & 49.6 & 35.8 & 29.9 & 32.8 & 27.1 \\
    Cascade RCNN~\cite{Cai_2019} & Swin-T~\cite{liu2021Swin} & & & 32.2 & 50.5 & 33.8 & 17.9 & 40.5 & 49.7 & 35.5 & 30.0 & 29.5 & 25.1 \\
    Cascade RCNN~\cite{Cai_2019} & ConvNeXt-T~\cite{liu2022convnet} & & & 34.8 & 53.3 & 36.7 & 20.6 & 42.8 & 52.5 & 37.7 & 32.1 & 36.1 & 28.4 \\
    \bottomrule
  \end{tabular}
  \caption{Evaluation of object detection using RGB images, with different pre-training and fine-tuning settings.
  }\label{tab:benchmark_rgb_resize}
\end{table*}

\subsection{Analysis}

With \nameofdata, we analyze the performances of various detectors for object detection with both sRGB and RAW images, as shown in \tabref{tab:benchmark_rgb_resize}. 
We evaluate some popular and milestone works, including multi-stage detectors~(Faster RCNN~\cite{Ren_2017}, Sparse RCNN~\cite{Sun_2021_CVPR}, and  Cascade RCNN~\cite{Cai_2019}), one-stage detectors (RetinaNet~\cite{lin2017focal} and GFocal~\cite{li2020generalized}), and transformer-based detectors (Deformable DETR~\cite{zhu2021deformable}). 

\myPara{sRGB object detection in adverse conditions.}
Cascade RCNN achieves superior performance among the evaluated methods, 
with 25.6\% AP and 27.3\% AP$_{\rm normal}$. 
However, the AP$_{\rm low}$, AP$_{\rm rain}$, and AP$_{\rm fog}$ are only 23.8\%, 24.7\%, and 20.4\%, showing that the adverse conditions bring more challenges. 
More advanced backbones like ConvNeXt and Swin-T 
can improve performance. 
For example, 
ConvNeXt-T outperforms ResNet by 9.7\% in AP$_{\rm normal}$, 
but with lower improvements on adverse conditions, 
\ie, 7.7\% AP$_{\rm low}$, 6.2\% AP$_{\rm rain}$, and 6.8\% AP$_{\rm fog}$. 
Such a gap shows the drawback of sRGB images in adverse conditions.

\myPara{RAW object detection in adverse conditions.}
It is inappropriate to adopt models pre-trained on sRGB images when fine-tuning on RAW images. 
For example, RAW-based Cascade RCNN only achieves 33.7\% AP when using sRGB pre-training, 
which is lower than 34.0\% of the sRGB-based method. 
This phenomenon is partially caused by the domain gap between sRGB and RAW. 
As shown in \tabref{tab:domain_shift}, 
the detector trained on one domain will be significantly degraded when testing on another, 
showing that 
the RAW and sRGB domain models cannot generalize well to each other. 

To overcome the domain gap, previous object detection methods in the RAW domain usually connect a neural image signal processor~(ISP) with detectors, where the ISP projects the images from RAW to the sRGB domain. 
In \tabref{tab:compare}, we evaluate two recently proposed methods designed for RAW object detection, RAOD~\cite{CVPR2023_ROD} and RAW-Adapter~\cite{raw_adapter}. 
Results show that neural ISP can stimulate the potential of RAW images. 
For example, RAOD achieves 34.4\% AP and outperforms the 34.0\% AP of the sRGB-based method. 
In particular, RAOD achieves greater improvements in adverse conditions than the normal condition (0.9\% AP$_{\rm low}$, 4.8\% AP$_{\rm rain}$, and 2.2\% AP$_{\rm fog}$ vs 0.3\% AP$_{\rm normal}$).
However, the neural ISP incurs extra computational costs and cannot fill the gap between RAW and sRGB domains, preventing the knowledge learned by the pre-trained model from being fully utilized. 

\section{RAW Pre-training}
\label{sec:raw_pre_training}

\subsection{Method}
Based on the above analysis, 
we aim to overcome the gap between sRGB pre-training and RAW fine-tuning 
by directly pre-training the models on RAW images, 
enabling us to achieve superior performances without requiring pre-processing modules like neural ISP. 

\myPara{Synthetic ImageNet-RAW.}
Visual pre-training~\cite{he2016deep,he2021masked} has made huge progress with 
the help of large-scale datasets like ImageNet-1K~\cite{russakovsky2015imagenet} that has over one million images. 
However, it is unrealistic to collect real RAW datasets of comparable size to large-scale sRGB datasets. 
Thus, we synthesize a 16-bit RAW dataset from the ImageNet-1K for RAW pre-training, 
using the unprocessing method~\cite{brooks2019unprocessing} that 
converts sRGB images to RAW images and simulates camera noise. 
We refer to the generated dataset as ImageNet-RAW. 
In the implementation, the unprocessed operation is inserted into the pipeline of data augmentations. 
Thus, 
we can randomly adjust the average brightness and simulated noise in each iteration 
so that models can generalize well across different conditions. 

\begin{table}
  \centering
  \setlength{\tabcolsep}{3.0mm}
  \begin{tabular}{ccccc}
    \toprule
    Training  & Evaluation  & AP & AP$_{\rm 50}$ & AP$_{\rm 75}$ \\
    \midrule
    \tRows{sRGB} & sRGB & 34.0 & 52.7 & 36.3 \\
    & RAW & 28.0 & 43.2 & 29.6 \\
    \midrule
    \tRows{RAW} & sRGB & 21.2 & 33.1 & 22.5 \\
    & RAW & 34.8 & 53.3 & 36.7 \\
    \bottomrule
  \end{tabular}
  \caption{The domain gap between sRGB and RAW.}
  \label{tab:domain_shift}
\end{table}

\begin{table*}
  \centering
  \setlength{\tabcolsep}{0.8mm}
  \begin{tabular}{llcccc|ccc|cccc}
    \toprule
    Method & Backbone & Neural ISP & AP & AP$_{\rm 50}$ & AP$_{\rm 75}$ & AP$_{\rm s}$ & AP$_{\rm m}$ & AP$_{\rm l}$ & AP$_{\rm normal}$ & AP$_{\rm low}$ & AP$_{\rm rain}$ & AP$_{\rm fog}$ \\
    \midrule
    \tRows{Baseline} & ConvNeXt-T~\cite{liu2022convnet} & \xmark & 33.4 & 51.8 & 35.3 & 19.4 & 41.1 & 50.7 & 36.8 & 31.0 & 30.2 & 27.0 \\
    & ResNet-18~\cite{he2016deep} & \xmark & 18.1 & 30.9 & 18.3 & 8.2 & 24.8 & 33.3 & 20.6 & 16.7 & 17.3 & 14.6 \\
    \midrule
    Gamma correction~\cite{Guo:04} & ConvNeXt-T~\cite{liu2022convnet} & \xmark & 33.7 & 52.0 & 35.9 & 18.6 & 41.7 & 51.3 & 36.8 & 31.3 & 31.3 & 27.2 \\
    RAOD~\cite{CVPR2023_ROD} & ConvNeXt-T~\cite{liu2022convnet} & \cmark & 34.4 & 52.9 & 35.9 & 19.5 & 42.9 & 52.2 & 37.1 & 31.9 & 35.0 & 29.2 \\
    RAW-Adapter~\cite{raw_adapter} & ResNet-18~\cite{he2016deep} & \cmark & 19.9 & 33.2 & 20.1 & 9.8 & 27.3 & 34.4 & 22.3 & 18.1 & 20.8 & 16.9 \\
    \midrule
    \tRows{Ours} & ConvNeXt-T~\cite{liu2022convnet} & \xmark & 34.8 & 53.3 & 36.7 & 20.6 & 42.8 & 52.5 & 37.7 & 32.1 & 36.1 & 28.4 \\ 
    & ResNet-18~\cite{he2016deep} & \xmark & 22.3 & 36.6 & 23.5 & 11.3 & 29.3 & 36.3 & 25.4 & 20.1 & 22.4 & 18.6 \\
    \bottomrule
  \end{tabular}
  \caption{Comparison with methods that adapt models pre-trained on sRGB domain to RAW images.}
  \label{tab:compare}
  \vspace{-20pt}
\end{table*}

\begin{figure}[t]
  \centering
  \begin{overpic}[width=0.9\linewidth]{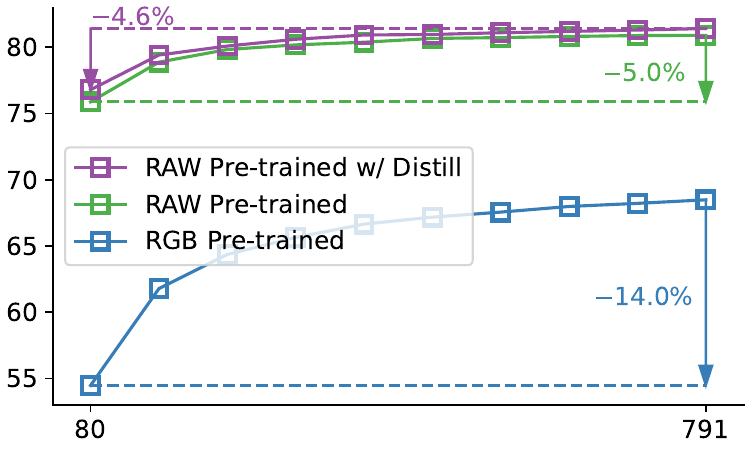}
    \put(-4, 18){\footnotesize \rotatebox{90}{Top-1 accuracy (\%)}}
    \put(37, 0){\footnotesize average brightness}
  \end{overpic}
  \caption{Top-1 accuracy on ImageNet-RAW when 
  synthesizing RAW images under different 
  average brightness. The 
  maximum average brightness for an image is $2^{16}$.}
  \label{fig:convnext_pretrain_accuracy}
  \vspace{-10pt}
\end{figure}

\myPara{RAW pre-training with cross-domain distillation.} 
By replacing the sRGB input with synthetic RAW images, we can pre-train models using the classification targets provided by the original ImageNet-1K dataset and keep the hyper-parameters consistent with sRGB pre-training. 
However, due to the noise in RAW images, it is more difficult for the models to learn high-quality representations in the RAW domain than in the sRGB domain. 
For example, the ConvNeXt-T pre-trained on ImageNet-1K can achieve 82.1\% Top-1 accuracy when testing on the sRGB domain. 
However, the model pre-trained on synthesized ImageNet-RAW only achieves 81.3\% Top-1 accuracy 
when testing on the RAW domain. 
To alleviate this problem, we propose using 
the knowledge distillation~\cite{hinton2015distilling,wei2022contrastive}. 
As cross-domain distillation, 
we take an off-the-shelf model pre-trained on the sRGB domain 
as the teacher 
to assist the pre-training on the RAW domain. 
The student shares the same architecture as the teacher 
for a fair comparison.
Specifically, a logit distillation with the Kullback-Leibler divergence loss and 
a feature distillation with the L1 loss are combined. 
The supplementary material shows the details and ablation studies.

\begin{figure}[t]
  \centering
  \begin{overpic}[width=0.9\linewidth]{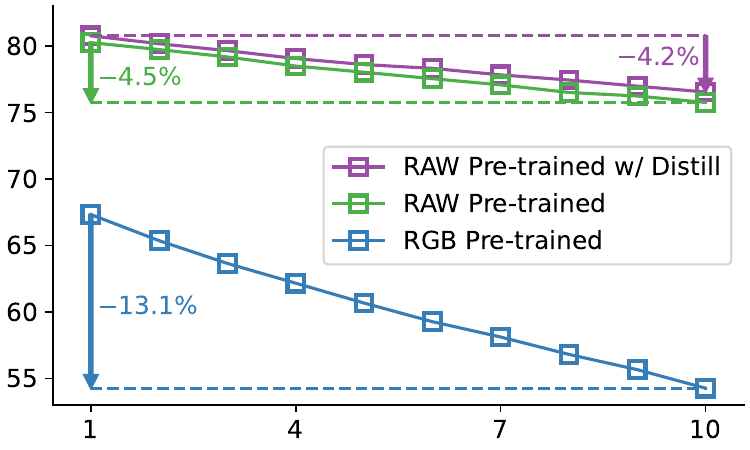}
    \put(-4, 18){\footnotesize \rotatebox{90}{Top-1 accuracy (\%)}}
    \put(46.0, -1){\footnotesize noise level}
  \end{overpic}
  \caption{Top-1 accuracy on ImageNet-RAW when adding different noise levels to synthesized RAW images. 
  Here, the noise level represents the standard deviation of the shot noises.}
  \label{fig:convnext_pretrain_accuracy_nose}
  \vspace{-10pt}
\end{figure}

\begin{table*}
  \centering
  \setlength{\tabcolsep}{0.7mm}
  \begin{tabular}{llccccc|ccc|cccc}
    \toprule
    Method & Backbone & Pre-Train & Fine-Tune & AP & AP$_{\rm 50}$ & AP$_{\rm 75}$ & AP$_{\rm s}$ & AP$_{\rm m}$ & AP$_{\rm l}$ & AP$_{\rm normal}$ & AP$_{\rm low}$ & AP$_{\rm rain}$ & AP$_{\rm fog}$ \\
    \midrule
    Cascade RCNN & Swin-T & \multirow{2}{*}{sRGB} & \multirow{2}{*}{sRGB} & 28.1 & 44.6 & 29.0 & 11.1 & 20.1 & 33.9 & 30.5 & 26.4 & 32.3 & 23.2 \\
    Cascade RCNN & ConvNeXt-T & & & 29.9 & 46.5 & 31.0 & 12.7 & 24.0 & 35.5 & 33.1 & 28.0 & 33.0 & 27.8 \\
    \midrule
    Cascade RCNN & Swin-T & \multirow{2}{*}{sRGB} & \multirow{2}{*}{RAW} & 29.2 & 46.2 & 30.2 & 10.9 & 19.8 & 35.1 & 31.0 & 27.8 & 32.3 & 24.6 \\
    Cascade RCNN & ConvNeXt-T & & & 29.7 & 46.9 & 30.6 & 11.5 & 22.2 & 35.4 & 32.3 & 27.8 & 33.1 & 27.0 \\
    \midrule
    Cascade RCNN & Swin-T & \multirow{2}{*}{RAW} & \multirow{2}{*}{RAW} & 29.8 & 47.0 & 30.9 & 11.4 & 21.7 & 35.4 & 31.4 & 28.1 & 32.9 & 27.3\\
    Cascade RCNN & ConvNeXt-T & & & 30.7 & 48.0 & 32.4 & 11.7 & 23.9 & 36.8 & 33.6 & 28.9 & 34.1 & 29.3 \\
    \bottomrule
  \end{tabular}
  \caption{Results of all-weather object detection using sliced RGB images.}
  \label{tab:benchmark_rgb_slice}
  \vspace{-5pt}
\end{table*}

\begin{table*}
  \centering
  \setlength{\tabcolsep}{1.0mm}
  \begin{tabular}{lcccccc|ccc|cccc}
    \toprule
    Method & Params~(M) & FPS & Epochs & AP & AP$_{\rm 50}$ & AP$_{\rm 75}$ & AP$_{\rm s}$ & AP$_{\rm m}$ & AP$_{\rm l}$ & AP$_{\rm normal}$ & AP$_{\rm low}$ & AP$_{\rm rain}$ & AP$_{\rm fog}$ \\
    \midrule
    YOLOX-Tiny~\cite{yolox2021} & 5.1 & 222.6 & 300 & 16.4 & 32.1 & 14.9 & 6.8 & 23.2 & 29.4 & 18.0 & 15.2 & 15.1 & 12.3 \\
    YOLOv6-n~\cite{jocheryolov82023} & 4.3 & 170.7 & 400 & 18.0 & 30.0 & 18.0 & 7.6 & 24.4 & 32.8 & 19.3 & 16.0 & 16.5 & 14.0 \\
    YOLOv8-n~\cite{jocheryolov82023} & 3.0 & 188.1 & 500 & 18.9 & 32.0 & 18.8 & 8.9 & 26.5 & 33.2 & 21.4 & 16.3 & 16.8 & 15.4 \\
    YOLOv8-n~\cite{jocheryolov82023}$^\dag$ & 3.1 & 57.6 & 500 & 19.7 & 32.8 & 19.9 & 9.4 & 27.0 & 32.9 & 21.8 & 16.9 & 18.9 & 14.9 \\
    YOLO-MS-XS~\cite{chen2023yoloms} & 4.5 & 113.0 & 300 & 24.7 & 40.0 & 25.1 & 12.1 & 33.4 & 41.4 & 28.2 & 22.4 & 21.2 & 19.7 \\
    \bottomrule
  \end{tabular}
  \caption{Evaluation of real-time object detection using down-sampled images. 
  The models are trained and evaluated with an input size of $1280\times 1280$. 
  $^\dag$ means using the trainable pre-processing module proposed by \cite{CVPR2023_ROD}. 
  Meanwhile, we measure the frames per second (FPS) of all models using an NVIDIA 3090 GPU.}
  \label{tab:yolo}
  \vspace{-15pt}
\end{table*}

\myPara{Enhanced robustness to adverse conditions.} 
RAW pre-training and cross-domain distillation 
enhance the robustness to different conditions. 
For one sRGB image, 
when converted to RAW at different epochs of pre-training, 
its brightness will be randomly adjusted, and 
random noise will be added. 
Meanwhile, 
the distillation provides consistent targets regardless of synthesized noise and brightness. 
Thus, the models are prompted to learn representations invariant to those conditions. 

\figref{fig:convnext_pretrain_accuracy} and \figref{fig:convnext_pretrain_accuracy_nose} 
verify the effects on robustness. 
By synthesizing the ImageNet-RAW validation set 
under different brightness and noise levels and evaluating the pre-trained models, 
we can observe that models pre-trained using distillation 
exhibit greater robustness to adverse conditions. 
For example, 
when reducing the brightness from 791 to 80, 
the performance degradation is 4.6\% Top-1 accuracy when distillation, 
lower than 5.0\% without distillation. 
Similarly, 
when increasing the noise level from 1 to 10, 
distillation exhibits a lower performance gap, 
\ie, 4.2\% vs 4.5\%. 
In addition, 
we show that sRGB pre-training has a significantly higher performance degradation of 
14.0\% and 13.1\% when adjusting the brightness and noise, respectively, 
showing that RAW pre-training effectively enhances the robustness.

\subsection{Experiments Results.} 
Through RAW pre-training and distillation, ConvNeXt-T achieves 81.8\% Top-1 accuracy on synthetic ImageNet-RAW. 
Although pre-training accuracy is still lower than sRGB pre-training~(82.1\%), 
the performance on real RAW object detection has significantly improved across various architectures, as shown in \tabref{tab:benchmark_rgb_resize}. 
For instance, RAW pre-training improves 
Cascade RCNN and ConvNeXt-T 
by 1.1\% AP than sRGB pre-training.  
In particular, the models achieve more significant improvements in adverse conditions, 
\ie, 0.8\% AP$_{\rm low}$, 4.8\% AP$_{\rm rain}$, and 1.2\% AP$_{\rm fog}$ vs 0.9\% AP$_{\rm normal}$. 
Compared to sRGB-based object detection, 
we improve by 0.8\% AP. 
These results demonstrate the advantages of our RAW pre-training.

\section{Experiments on Sliced \nameofdata}

The experiments in \secref{sec:benchmark} and \secref{sec:raw_pre_training} use the down-sampling setting. 
\tabref{tab:benchmark_rgb_slice} further lists the results of the slicing setting, where the hyper-parameters follow the down-sampling experiments, 
except that we fine-tune models for 12 epochs. 
The experiments with ConvNeXt show the same trends as those in \tabref{tab:benchmark_rgb_resize}. 
Adapting models pre-trained on the sRGB domain to RAW object detection degrades the AP by 0.2\% compared to sRGB object detection, and RAW pre-training improves the performance by 0.8\%. 
Differently, with the Swin transformer, RAW object detection outperforms sRGB object detection even when using sRGB pre-training. 
It may be because the sliced images contain more visual information, allowing the model to converge well compared to down-sampled images. 
Meanwhile, RAW pre-training further improves performance by 0.6\% AP, especially in adverse conditions. 
Overall, these results further illustrate the effectiveness of RAW pre-training. 

\section{Experiments on Real-Time Object Detection}
Besides the regular detectors in \tabref{tab:benchmark_rgb_resize}, 
we also evaluate real-time object detection of YOLO-Series. 
As shown in \tabref{tab:yolo}, 
YOLO-MS-XS~\cite{chen2023yoloms} and YOLO-v8-n~\cite{jocheryolov82023} 
achieve 24.7\% and 18.9\% AP with a high FPS and parameters less than 5M. 
However, adverse conditions have lower performance, 
such as 16.3\% AP$_{\rm low}$, 16.8\% AP$_{\rm rain}$, and 15.4\% AP$_{\rm low}$ of YOLO-v8-n. 
Some methods have tried to boost the performance via a trainable ISP. 
Here, 
we take the recently proposed method~\cite{CVPR2023_ROD} as an example for analysis. 
In \tabref{tab:yolo}, 
integrating~\cite{CVPR2023_ROD} with YOLO-v8-n 
improves the performance to 16.9\% AP$_{\rm low}$ and 18.9\% AP$_{\rm rain}$. 
However, the FPS is significantly reduced, 
destroying the real-time property of the detector. 
In summary, the proposed \nameofdata~provides a new foundation 
to push real-time RAW object detection development.

\section{Conclusion}
\label{sec:conclusions}

This paper introduces \nameofdata, a challenging dataset for RAW-based object detection across diverse and adverse conditions.
Compared to traditional sRGB datasets, \nameofdata{} offers diverse RAW images that retain essential visual information for object detection 
in complex light and weather conditions. 
Based on \nameofdata, we evaluate existing methods of RAW object detection. 
Meanwhile, 
we utilize a cross-domain knowledge distillation 
to directly pre-train models on the RAW domain, 
solving the domain gap between sRGB pre-training and RAW fine-tuning. 
In this way, 
we improve the performance, particularly in adverse conditions, 
without relying on extra pre-processing modules. 
Our dataset is a benchmark for evaluating detection methods and a foundation for developing detection methods that generalize well across varied conditions. 
Our insights highlight the potential of RAW pre-training to advance real-world object detection and encourage further research on exploiting RAW images for challenging environments.

{
\small
\bibliographystyle{ieeenat_fullname}
\bibliography{main}

\begin{thebibliography}{38}
\providecommand{\natexlab}[1]{#1}
\providecommand{\url}[1]{\texttt{#1}}
\expandafter\ifx\csname urlstyle\endcsname\relax
  \providecommand{\doi}[1]{doi: #1}\else
  \providecommand{\doi}{doi: \begingroup \urlstyle{rm}\Url}\fi

\bibitem[Bijelic et~al.(2020)Bijelic, Gruber, Mannan, Kraus, Ritter, Dietmayer,
  and Heide]{Bijelic_2020_CVPR}
Mario Bijelic, Tobias Gruber, Fahim Mannan, Florian Kraus, Werner Ritter, Klaus
  Dietmayer, and Felix Heide.
\newblock Seeing through fog without seeing fog: Deep multimodal sensor fusion
  in unseen adverse weather.
\newblock In \emph{CVPR}, 2020.

\bibitem[Brooks et~al.(2019)Brooks, Mildenhall, Xue, Chen, Sharlet, and
  Barron]{brooks2019unprocessing}
Tim Brooks, Ben Mildenhall, Tianfan Xue, Jiawen Chen, Dillon Sharlet, and
  Jonathan~T Barron.
\newblock Unprocessing images for learned raw denoising.
\newblock In \emph{CVPR}, 2019.

\bibitem[Cai and Vasconcelos(2019)]{Cai_2019}
Zhaowei Cai and Nuno Vasconcelos.
\newblock Cascade r-cnn: High quality object detection and instance
  segmentation.
\newblock \emph{IEEE TPAMI}, 2019.

\bibitem[Chen et~al.(2019)Chen, Wang, Pang, Cao, Xiong, Li, Sun, Feng, Liu, Xu,
  Zhang, Cheng, Zhu, Cheng, Zhao, Li, Lu, Zhu, Wu, Dai, Wang, Shi, Ouyang, Loy,
  and Lin]{mmdetection}
Kai Chen, Jiaqi Wang, Jiangmiao Pang, Yuhang Cao, Yu Xiong, Xiaoxiao Li,
  Shuyang Sun, Wansen Feng, Ziwei Liu, Jiarui Xu, Zheng Zhang, Dazhi Cheng,
  Chenchen Zhu, Tianheng Cheng, Qijie Zhao, Buyu Li, Xin Lu, Rui Zhu, Yue Wu,
  Jifeng Dai, Jingdong Wang, Jianping Shi, Wanli Ouyang, Chen~Change Loy, and
  Dahua Lin.
\newblock {MMDetection}: Open mmlab detection toolbox and benchmark.
\newblock \emph{arXiv preprint arXiv:1906.07155}, 2019.

\bibitem[Chen et~al.(2023{\natexlab{a}})Chen, Fu, Wei, Zheng, and
  Heide]{2023lis}
Linwei Chen, Ying Fu, Kaixuan Wei, Dezhi Zheng, and Felix Heide.
\newblock Instance segmentation in the dark.
\newblock \emph{International Journal of Computer Vision}, 131\penalty0
  (8):\penalty0 2198--2218, 2023{\natexlab{a}}.

\bibitem[Chen et~al.(2023{\natexlab{b}})Chen, Yuan, Wu, Wang, Hou, and
  Cheng]{chen2023yoloms}
Yuming Chen, Xinbin Yuan, Ruiqi Wu, Jiabao Wang, Qibin Hou, and Ming-Ming
  Cheng.
\newblock Yolo-ms: Rethinking multi-scale representation learning for real-time
  object detection.
\newblock \emph{arXiv preprint arXiv:2308.05480}, 2023{\natexlab{b}}.

\bibitem[Cheng et~al.(2023)Cheng, Yuan, Yao, Yan, Zeng, Xie, and Han]{SODA}
Gong Cheng, Xiang Yuan, Xiwen Yao, Kebing Yan, Qinghua Zeng, Xingxing Xie, and
  Junwei Han.
\newblock Towards large-scale small object detection: Survey and benchmarks.
\newblock \emph{IEEE TPAMI}, 45\penalty0 (11):\penalty0 13467--13488, 2023.

\bibitem[Cui and Harada(2024)]{raw_adapter}
Ziteng Cui and Tatsuya Harada.
\newblock Raw-adapter: Adapting pretrained visual model to camera raw images.
\newblock In \emph{ECCV}, 2024.

\bibitem[Diamond et~al.(2021)Diamond, Sitzmann, Julca-Aguilar, Boyd, Wetzstein,
  and Heide]{Dirtypixel}
Steven Diamond, Vincent Sitzmann, Frank Julca-Aguilar, Stephen Boyd, Gordon
  Wetzstein, and Felix Heide.
\newblock Dirty pixels: Towards end-to-end image processing and perception.
\newblock \emph{ACM Trans. Graph.}, 40\penalty0 (3), 2021.

\bibitem[Everingham et~al.(2009)Everingham, Gool, Williams, Winn, and
  Zisserman]{Everingham2009ThePV}
M. Everingham, L. Gool, Christopher K.~I. Williams, J. Winn, and Andrew
  Zisserman.
\newblock The pascal visual object classes (voc) challenge.
\newblock \emph{IJCV}, 88:\penalty0 303--338, 2009.

\bibitem[Ge et~al.(2021)Ge, Liu, Wang, Li, and Sun]{yolox2021}
Zheng Ge, Songtao Liu, Feng Wang, Zeming Li, and Jian Sun.
\newblock {YOLOX}: Exceeding yolo series in 2021.
\newblock \emph{arXiv preprint arXiv:2107.08430}, 2021.

\bibitem[Guo et~al.(2004)Guo, He, and Chen]{Guo:04}
Hongwei Guo, Haitao He, and Mingyi Chen.
\newblock Gamma correction for digital fringe projection profilometry.
\newblock \emph{Appl. Opt.}, 43\penalty0 (14):\penalty0 2906--2914, 2004.

\bibitem[Guo et~al.(2024)Guo, Luo, and Wu]{Guo_2024_CVPR}
Yanhui Guo, Fangzhou Luo, and Xiaolin Wu.
\newblock Learning degradation-independent representations for camera isp
  pipelines.
\newblock In \emph{CVPR}, 2024.

\bibitem[He et~al.(2016)He, Zhang, Ren, and Sun]{he2016deep}
Kaiming He, Xiangyu Zhang, Shaoqing Ren, and Jian Sun.
\newblock Deep residual learning for image recognition.
\newblock In \emph{CVPR}, 2016.

\bibitem[He et~al.(2022)He, Chen, Xie, Li, Doll\'ar, and
  Girshick]{he2021masked}
Kaiming He, Xinlei Chen, Saining Xie, Yanghao Li, Piotr Doll\'ar, and Ross
  Girshick.
\newblock Masked autoencoders are scalable vision learners.
\newblock In \emph{CVPR}, 2022.

\bibitem[Hinton(2015)]{hinton2015distilling}
Geoffrey Hinton.
\newblock Distilling the knowledge in a neural network.
\newblock \emph{arXiv preprint arXiv:1503.02531}, 2015.

\bibitem[Hong et~al.(2021)Hong, Wei, Chen, and Fu]{Hong2021Crafting}
Yang Hong, Kaixuan Wei, Linwei Chen, and Ying Fu.
\newblock Crafting object detection in very low light.
\newblock In \emph{BMVC}, 2021.

\bibitem[Jocher et~al.(2023)Jocher, Chaurasia, and Qiu]{jocheryolov82023}
Glenn Jocher, Ayush Chaurasia, and Jing Qiu.
\newblock {YOLO by Ultralytics}, 2023.

\bibitem[Li et~al.(2020)Li, Wang, Wu, Chen, Hu, Li, Tang, and
  Yang]{li2020generalized}
Xiang Li, Wenhai Wang, Lijun Wu, Shuo Chen, Xiaolin Hu, Jun Li, Jinhui Tang,
  and Jian Yang.
\newblock Generalized focal loss: Learning qualified and distributed bounding
  boxes for dense object detection.
\newblock In \emph{NeurIPS}, 2020.

\bibitem[Li et~al.(2024)Li, Lu, Zhang, Feng, Asif, and Ma]{10415533}
Zhihao Li, Ming Lu, Xu Zhang, Xin Feng, M.~Salman Asif, and Zhan Ma.
\newblock Efficient visual computing with camera raw snapshots.
\newblock \emph{IEEE TPAMI}, 46\penalty0 (7):\penalty0 4684--4701, 2024.

\bibitem[Lin et~al.(2014)Lin, Maire, Belongie, Hays, Perona, Ramanan,
  Doll{\'a}r, and Zitnick]{lin2015microsoft}
Tsung-Yi Lin, Michael Maire, Serge Belongie, James Hays, Pietro Perona, Deva
  Ramanan, Piotr Doll{\'a}r, and C~Lawrence Zitnick.
\newblock Microsoft coco: Common objects in context.
\newblock In \emph{ECCV}, 2014.

\bibitem[Lin et~al.(2017)Lin, Goyal, Girshick, He, and
  Doll{\'a}r]{lin2017focal}
Tsung-Yi Lin, Priya Goyal, Ross Girshick, Kaiming He, and Piotr Doll{\'a}r.
\newblock Focal loss for dense object detection.
\newblock In \emph{ICCV}, 2017.

\bibitem[Liu et~al.(2021)Liu, Lin, Cao, Hu, Wei, Zhang, Lin, and
  Guo]{liu2021Swin}
Ze Liu, Yutong Lin, Yue Cao, Han Hu, Yixuan Wei, Zheng Zhang, Stephen Lin, and
  Baining Guo.
\newblock Swin transformer: Hierarchical vision transformer using shifted
  windows.
\newblock In \emph{ICCV}, 2021.

\bibitem[Liu et~al.(2022)Liu, Mao, Wu, Feichtenhofer, Darrell, and
  Xie]{liu2022convnet}
Zhuang Liu, Hanzi Mao, Chao-Yuan Wu, Christoph Feichtenhofer, Trevor Darrell,
  and Saining Xie.
\newblock A convnet for the 2020s.
\newblock In \emph{CVPR}, 2022.

\bibitem[Mosleh et~al.(2020)Mosleh, Sharma, Onzon, Mannan, Robidoux, and
  Heide]{Mosleh_2020_CVPR}
Ali Mosleh, Avinash Sharma, Emmanuel Onzon, Fahim Mannan, Nicolas Robidoux, and
  Felix Heide.
\newblock Hardware-in-the-loop end-to-end optimization of camera image
  processing pipelines.
\newblock In \emph{CVPR}, 2020.

\bibitem[Omid-Zohoor et~al.(2014)Omid-Zohoor, Ta, and
  Murmann]{omid2014pascalraw}
Alex Omid-Zohoor, David Ta, and Boris Murmann.
\newblock Pascalraw: raw image database for object detection.
\newblock \emph{Stanford Digital Repository}, 2014.

\bibitem[Qin et~al.(2022)Qin, Han, Wang, Zhang, Li, Li, and
  Hu]{qin2022attention}
Haina Qin, Longfei Han, Juan Wang, Congxuan Zhang, Yanwei Li, Bing Li, and
  Weiming Hu.
\newblock Attention-aware learning for hyperparameter prediction in image
  processing pipelines.
\newblock In \emph{ECCV}, 2022.

\bibitem[Ren et~al.(2017)Ren, He, Girshick, and Sun]{Ren_2017}
Shaoqing Ren, Kaiming He, Ross Girshick, and Jian Sun.
\newblock Faster r-cnn: Towards real-time object detection with region proposal
  networks.
\newblock \emph{IEEE TPAMI}, 2017.

\bibitem[Russakovsky et~al.(2015)Russakovsky, Deng, Su, Krause, Satheesh, Ma,
  Huang, Karpathy, Khosla, Bernstein, et~al.]{russakovsky2015imagenet}
Olga Russakovsky, Jia Deng, Hao Su, Jonathan Krause, Sanjeev Satheesh, Sean Ma,
  Zhiheng Huang, Andrej Karpathy, Aditya Khosla, Michael Bernstein, et~al.
\newblock Imagenet large scale visual recognition challenge.
\newblock \emph{IJCV}, 2015.

\bibitem[Sun et~al.(2021)Sun, Zhang, Jiang, Kong, Xu, Zhan, Tomizuka, Li, Yuan,
  Wang, and Luo]{Sun_2021_CVPR}
Peize Sun, Rufeng Zhang, Yi Jiang, Tao Kong, Chenfeng Xu, Wei Zhan, Masayoshi
  Tomizuka, Lei Li, Zehuan Yuan, Changhu Wang, and Ping Luo.
\newblock Sparse r-cnn: End-to-end object detection with learnable proposals.
\newblock In \emph{CVPR}, 2021.

\bibitem[Wang et~al.(2024)Wang, Xu, Zhang, Xue, and Gu]{wang2024adaptiveisp}
Yujin Wang, Tianyi Xu, Fan Zhang, Tianfan Xue, and Jinwei Gu.
\newblock Adaptiveisp: Learning an adaptive image signal processor for object
  detection.
\newblock In \emph{NeurIPS}, 2024.

\bibitem[Wei et~al.(2022)Wei, Hu, Xie, Zhang, Cao, Bao, Chen, and
  Guo]{wei2022contrastive}
Yixuan Wei, Han Hu, Zhenda Xie, Zheng Zhang, Yue Cao, Jianmin Bao, Dong Chen,
  and Baining Guo.
\newblock Contrastive learning rivals masked image modeling in fine-tuning via
  feature distillation.
\newblock \emph{arXiv preprint arXiv:2205.14141}, 2022.

\bibitem[Wu et~al.(2019)Wu, Isikdogan, Rao, Nayak, Gerasimow, Sutic, Ain-kedem,
  and Michael]{8803607}
Chyuan-Tyng Wu, Leo~F. Isikdogan, Sushma Rao, Bhavin Nayak, Timo Gerasimow,
  Aleksandar Sutic, Liron Ain-kedem, and Gilad Michael.
\newblock Visionisp: Repurposing the image signal processor for computer vision
  applications.
\newblock In \emph{ICIP}, 2019.

\bibitem[Xu et~al.(2023)Xu, Chen, Peng, Li, Huang, Song, Yan, and
  Xiong]{CVPR2023_ROD}
Ruikang Xu, Chang Chen, Jingyang Peng, Cheng Li, Yibin Huang, Fenglong Song,
  Youliang Yan, and Zhiwei Xiong.
\newblock Toward raw object detection: A new benchmark and a new model.
\newblock In \emph{CVPR}, 2023.

\bibitem[Yoshimura et~al.(2023)Yoshimura, Otsuka, Irie, and
  Ohashi]{Yoshimura_2023_ICCV}
Masakazu Yoshimura, Junji Otsuka, Atsushi Irie, and Takeshi Ohashi.
\newblock Dynamicisp: Dynamically controlled image signal processor for image
  recognition.
\newblock In \emph{ICCV}, 2023.

\bibitem[Yu et~al.(2021)Yu, Li, Peng, Loy, and Gu]{Yu_2021_ICCV}
Ke Yu, Zexian Li, Yue Peng, Chen~Change Loy, and Jinwei Gu.
\newblock Reconfigisp: Reconfigurable camera image processing pipeline.
\newblock In \emph{ICCV}, 2021.

\bibitem[Zhang et~al.(2023)Zhang, Guo, Yang, Zhang, Xie, Suo, and
  Dai]{zhang2023darkvision}
Bo Zhang, Yuchen Guo, Runzhao Yang, Zhihong Zhang, Jiayi Xie, Jinli Suo, and
  Qionghai Dai.
\newblock Darkvision: a benchmark for low-light image/video perception.
\newblock \emph{arXiv preprint arXiv:2301.06269}, 2023.

\bibitem[Zhu et~al.(2021)Zhu, Su, Lu, Li, Wang, and Dai]{zhu2021deformable}
Xizhou Zhu, Weijie Su, Lewei Lu, Bin Li, Xiaogang Wang, and Jifeng Dai.
\newblock Deformable detr: Deformable transformers for end-to-end object
  detection.
\newblock In \emph{ICLR}, 2021.

\end{thebibliography}
}

\clearpage
\setcounter{page}{1}
\maketitlesupplementary

\section{Statistics of \nameofdata}

\begin{figure*}[t]
  \centering
  \begin{overpic}[width=0.9\linewidth]{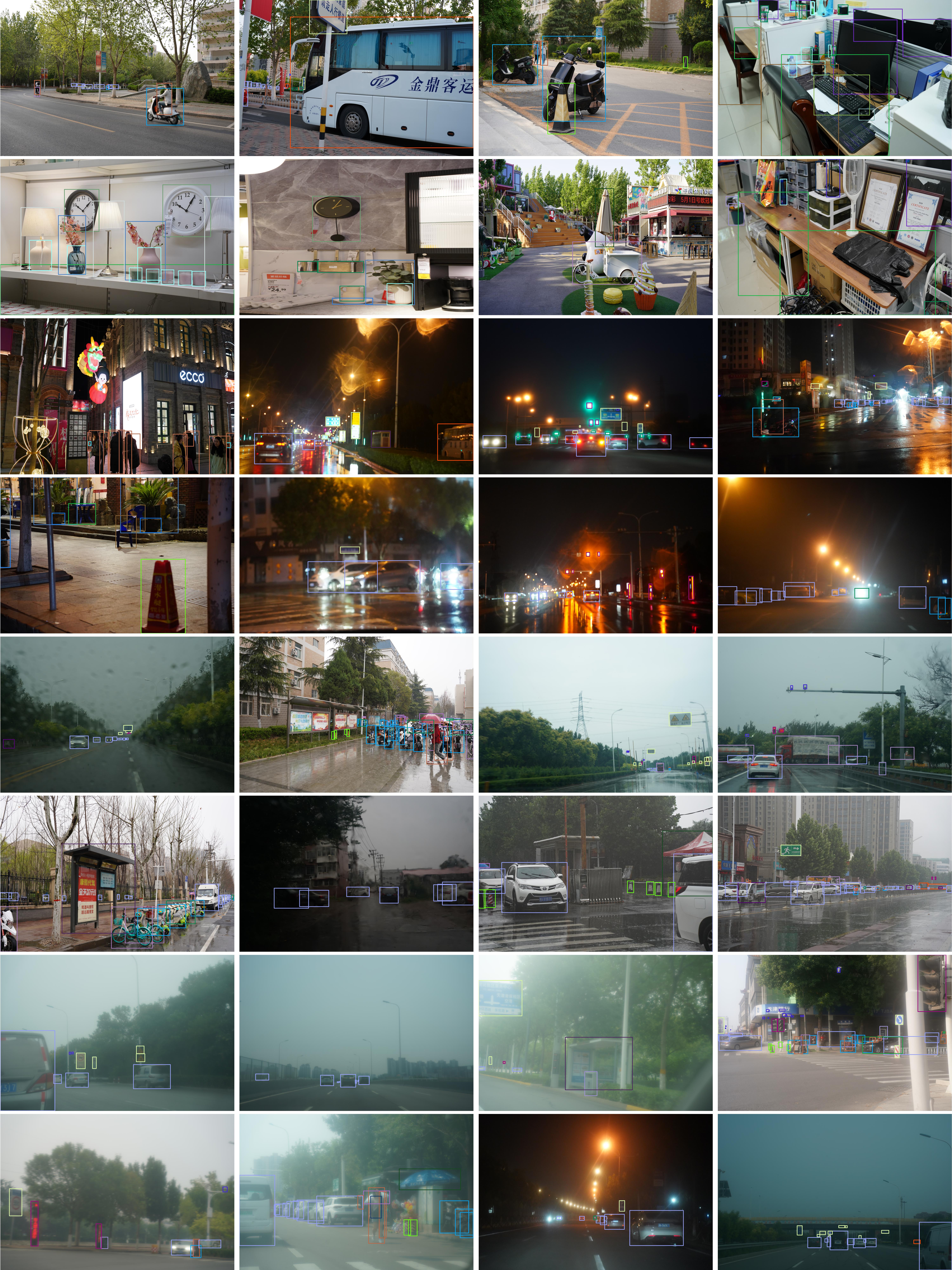}
    \put(-2, 79){\rotatebox{90}{Clear weather and Daylight}}
    \put(-2, 59){\rotatebox{90}{Low-light}}
    \put(-2, 36){\rotatebox{90}{Rain}}
    \put(-2, 11){\rotatebox{90}{Fog}}
  \end{overpic}
  \caption{Example of the images in the \nameofdata. 
  }\label{fig:more_dataset_example}
\end{figure*}

\myPara{More examples from \nameofdata.} 
We collect images under 9 conditions, as shown in Tab.~2 of the main paper. 
\tabref{tab:example_each_condition} 
shows \textbf{a specific example for each condition} for a better understanding.
Furthermore, 
\figref{fig:more_dataset_example} shows \textbf{more examples of our \nameofdata~dataset} and the annotated bounding boxes, 
showing the diversity of the \nameofdata. 

\begin{figure*}[t]
  \centering
  \begin{overpic}[width=0.9\linewidth]{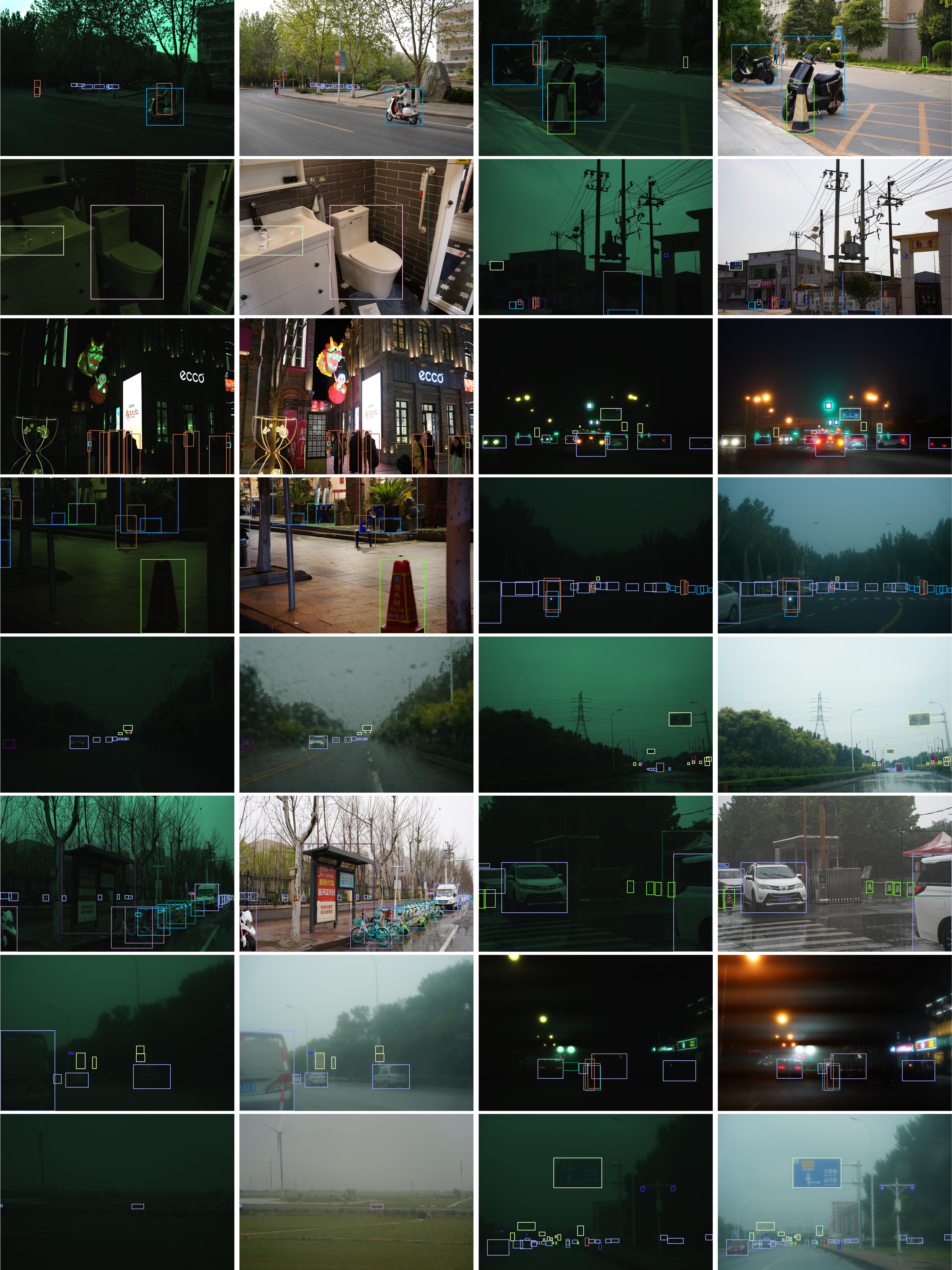}
    \put(08, -2){RAW}
    \put(26, -2){sRGB}
    \put(45, -2){RAW}
    \put(64, -2){sRGB}
  \end{overpic}
  \vspace{10pt}
  \caption{Example of the RAW images in the \nameofdata. 
  }\label{fig:dataset_example_raw}
\end{figure*}

\myPara{Visualization of RAW images.} 
To better show the domain gap between RAW and sRGB images, 
we show some RAW images and the corresponding sRGB images in 
\figref{fig:dataset_example_raw}. 
Here, 
the RAW images are visualized by demosaicing and normalizing them to the range of [0, 255],  
without any other preprocessing operations.

\myPara{Annotated categories.} 
\tabref{tab:annotated_categories} lists the categories in \nameofdata~and 
the corresponding number of instances. 
We sort the categories according to the number of instances. 

\begin{figure*}[t]
  \centering
  \begin{subfigure}[t]{0.26\textwidth}
    \begin{overpic}[width=\linewidth]{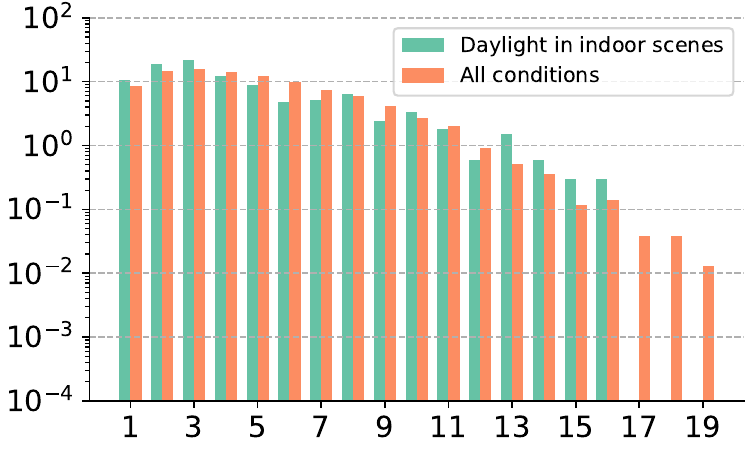}
      \put(-3, 9.5){\rotatebox{90}{\scriptsize Percent of images (\%)}}
    \end{overpic}
    \caption{Daylight in indoor scenes.}
    \label{fig:category_per_image_normal_light_indoor}
  \end{subfigure}
  \begin{subfigure}[t]{0.26\textwidth}
    \begin{overpic}[width=\linewidth]{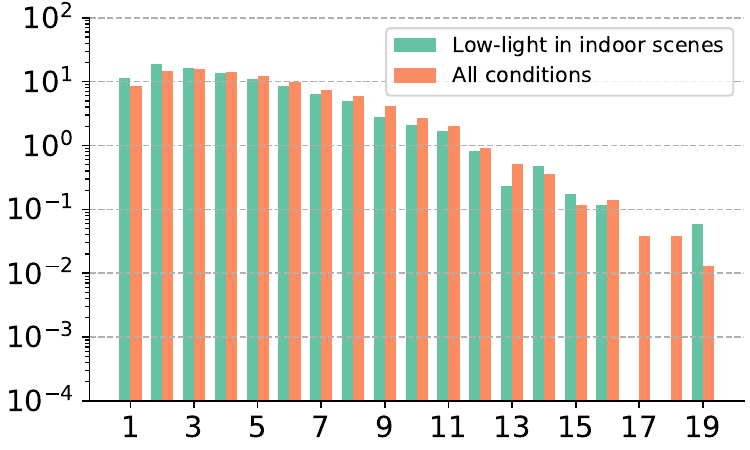}
      \put(-3, 9.5){\rotatebox{90}{\scriptsize Percent of images (\%)}}
    \end{overpic}
    \caption{Low-light in indoor scenes.}
    \label{fig:category_per_image_low_light_indoor}
  \end{subfigure}
  \begin{subfigure}[t]{0.26\textwidth}
    \begin{overpic}[width=\linewidth]{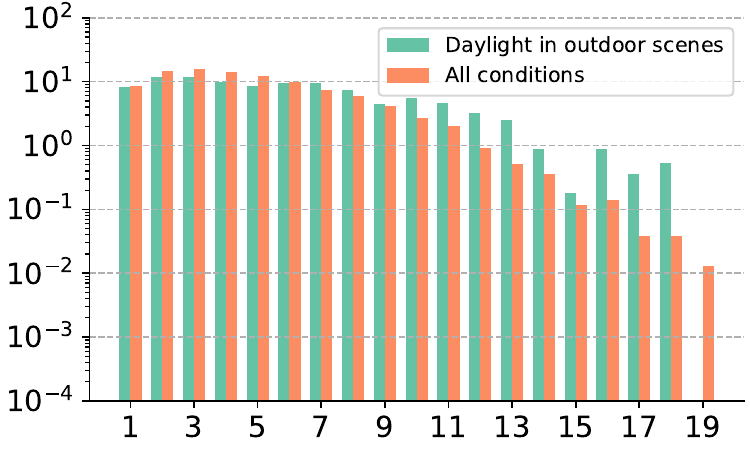}
      \put(-3, 9.5){\rotatebox{90}{\scriptsize Percent of images (\%)}}
    \end{overpic}
    \caption{Daylight in outdoor scenes.}
    \label{fig:category_per_image_normal_light}
  \end{subfigure}
  \begin{subfigure}[t]{0.26\textwidth}
    \begin{overpic}[width=\linewidth]{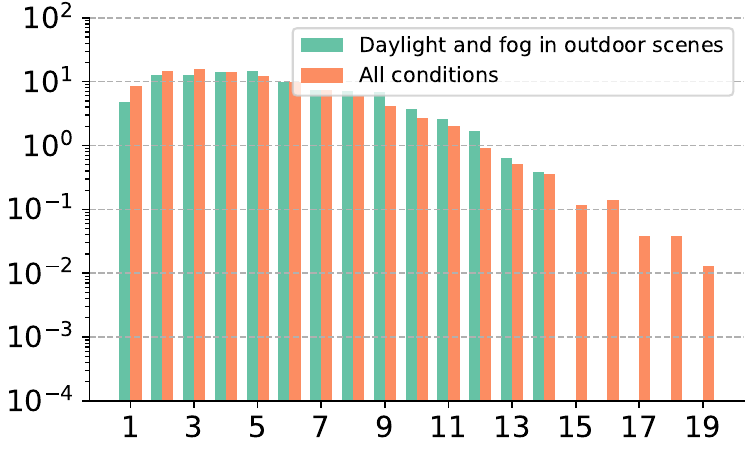}
      \put(-3, 9.5){\rotatebox{90}{\scriptsize Percent of images (\%)}}
    \end{overpic}
    \caption{Daylight and fog in outdoor scenes.}
    \label{fig:category_per_image_fog}
  \end{subfigure}
  \begin{subfigure}[t]{0.26\textwidth}
    \begin{overpic}[width=\linewidth]{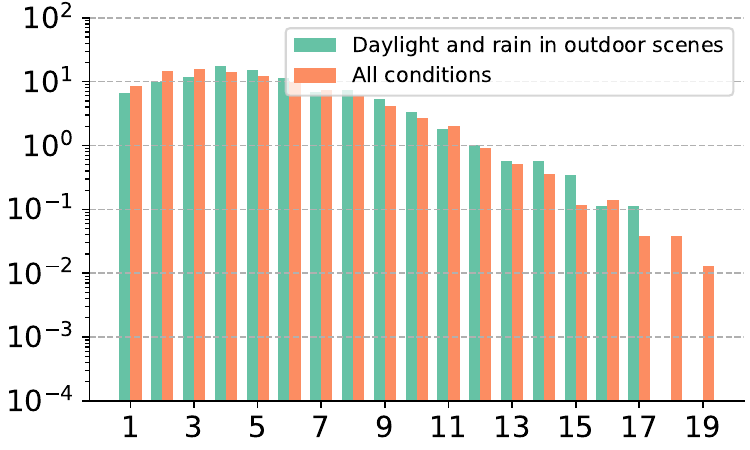}
      \put(-3, 9.5){\rotatebox{90}{\scriptsize Percent of images (\%)}}
    \end{overpic}
    \caption{Daylight and rain in outdoor scenes.}
    \label{fig:category_per_image_rain}
  \end{subfigure}
  \begin{subfigure}[t]{0.26\textwidth}
    \begin{overpic}[width=\linewidth]{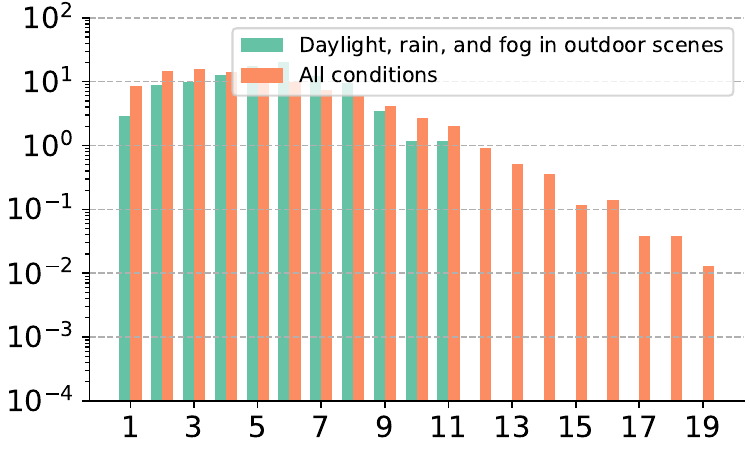}
      \put(-3, 9.5){\rotatebox{90}{\scriptsize Percent of images (\%)}}
    \end{overpic}
    \caption{Daylight, rain, and fog in outdoor scenes.}
    \label{fig:category_per_image_rain_fog}
  \end{subfigure}
  \begin{subfigure}[t]{0.26\textwidth}
    \begin{overpic}[width=\linewidth]{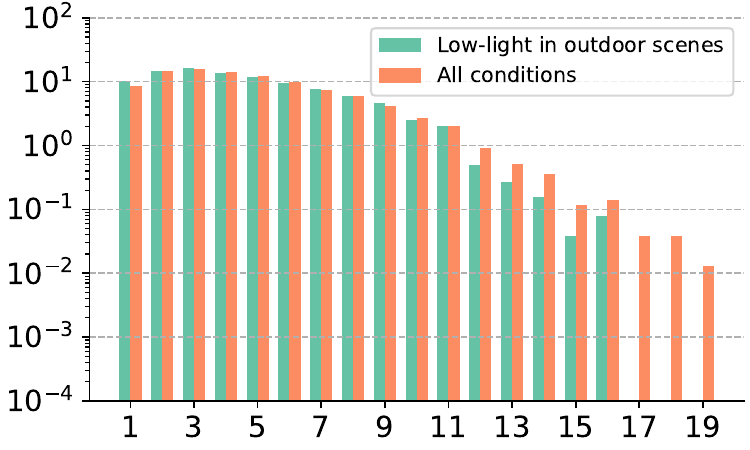}
      \put(-3, 9.5){\rotatebox{90}{\scriptsize Percent of images (\%)}}
    \end{overpic}
    \caption{Low-light in outdoor scenes.}
    \label{fig:category_per_image_low_light}
  \end{subfigure}
  \begin{subfigure}[t]{0.26\textwidth}
    \begin{overpic}[width=\linewidth]{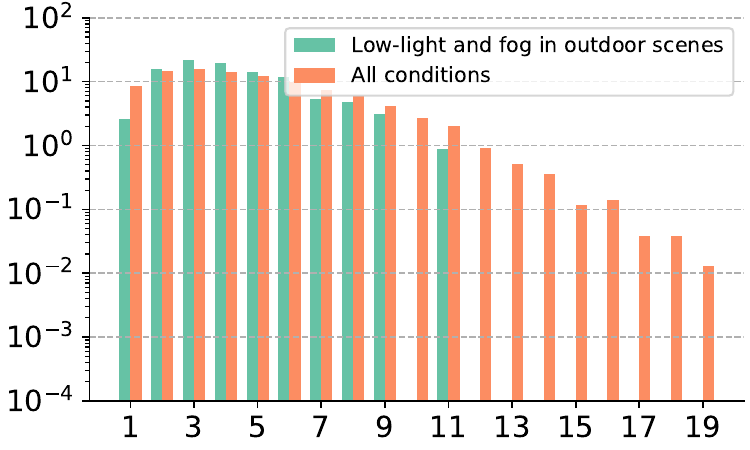}
      \put(-3, 9.5){\rotatebox{90}{\scriptsize Percent of images (\%)}}
    \end{overpic}
    \caption{Low-light and fog in outdoor scenes.}
    \label{fig:category_per_image_low_light_fog}
  \end{subfigure}
  \begin{subfigure}[t]{0.26\textwidth}
    \begin{overpic}[width=\linewidth]{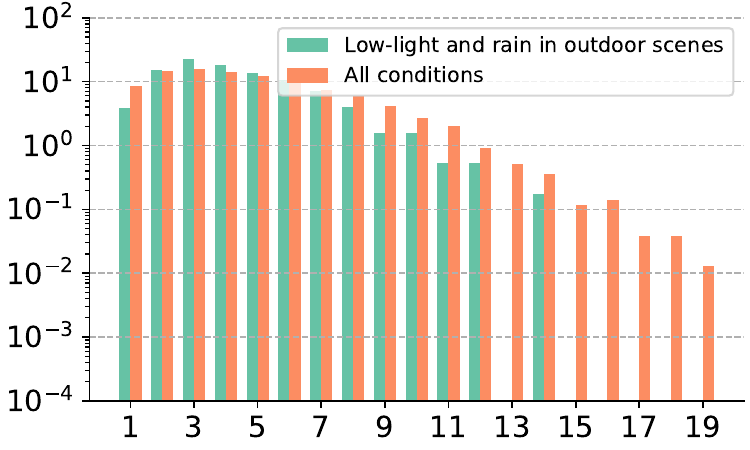}
      \put(-3, 9.5){\rotatebox{90}{\scriptsize Percent of images (\%)}}
    \end{overpic}
    \caption{Low-light and rain in outdoor scenes.}
    \label{fig:category_per_image_low_light_rain}
  \end{subfigure}
  \caption{Distribution of the number of categories in images of each condition. 
  The horizontal axis represents the number of categories.}
  \label{fig:category_per_iage_per_condition}
\end{figure*}

\myPara{Categories per image.}
\figref{fig:category_per_iage_per_condition} 
shows the distribution of the number of categories in the images. 
For each condition, 
the maximum number of categories exceeds 10. 
More than half of the conditions have a maximum number that exceeds 15. 
Especially, 
the maximum number for the indoor low-light condition is 19.

\begin{figure*}
  \centering
  \begin{subfigure}[t]{0.26\textwidth}
    \begin{overpic}[width=\linewidth]{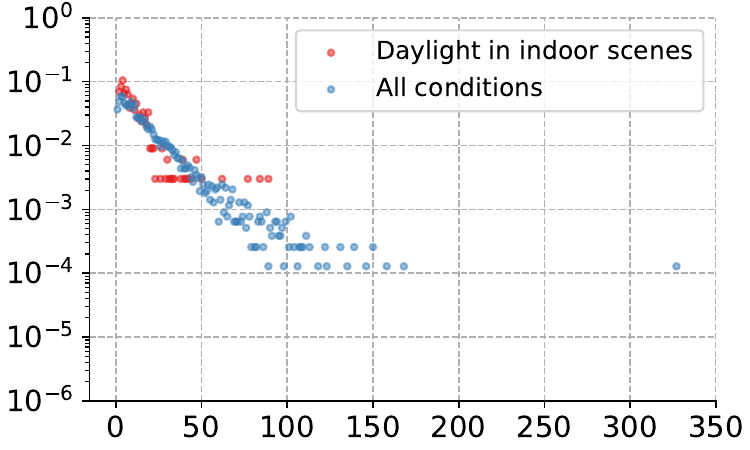}
      \put(-4, 9.5){\rotatebox{90}{\scriptsize Percent of images (\%)}}
    \end{overpic}
    \caption{Daylight in indoor scenes.}
    \label{fig:instance_per_image_normal_light_indoor}
  \end{subfigure}
  \begin{subfigure}[t]{0.26\textwidth}
    \begin{overpic}[width=\linewidth]{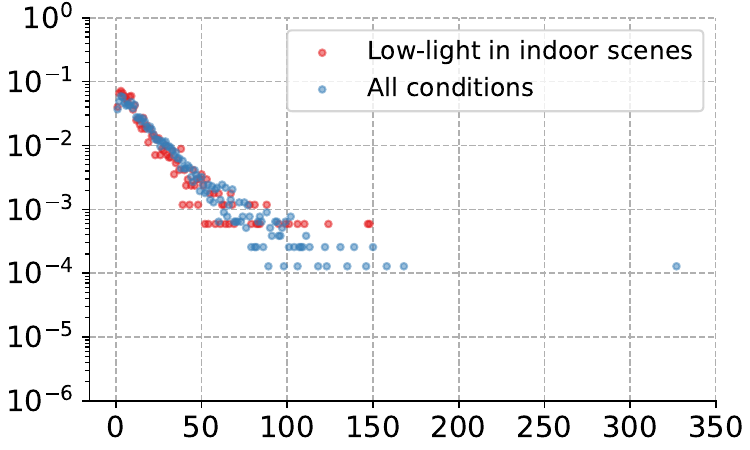}
      \put(-4, 9.5){\rotatebox{90}{\scriptsize Percent of images (\%)}}
    \end{overpic}
    \caption{Low-light in indoor scenes.}
    \label{fig:instance_per_image_low_light_indoor}
  \end{subfigure}
  \begin{subfigure}[t]{0.26\textwidth}
    \begin{overpic}[width=\linewidth]{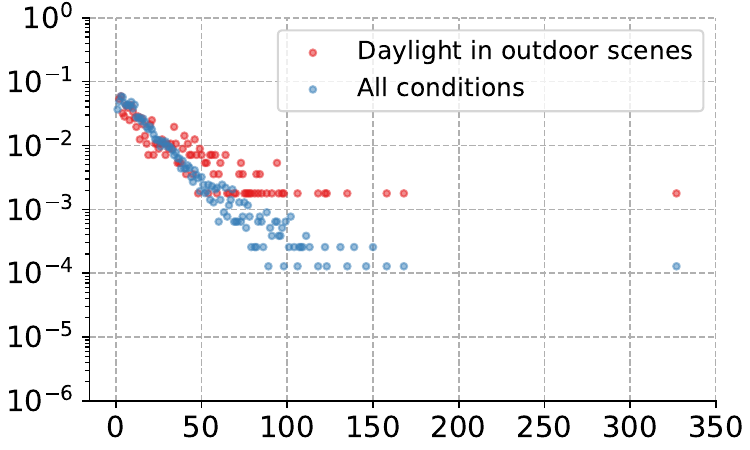}
      \put(-4, 9.5){\rotatebox{90}{\scriptsize Percent of images (\%)}}
    \end{overpic}
    \caption{Daylight in outdoor scenes.}
    \label{fig:instance_per_image_normal_light}
  \end{subfigure}
  \begin{subfigure}[t]{0.26\textwidth}
    \begin{overpic}[width=\linewidth]{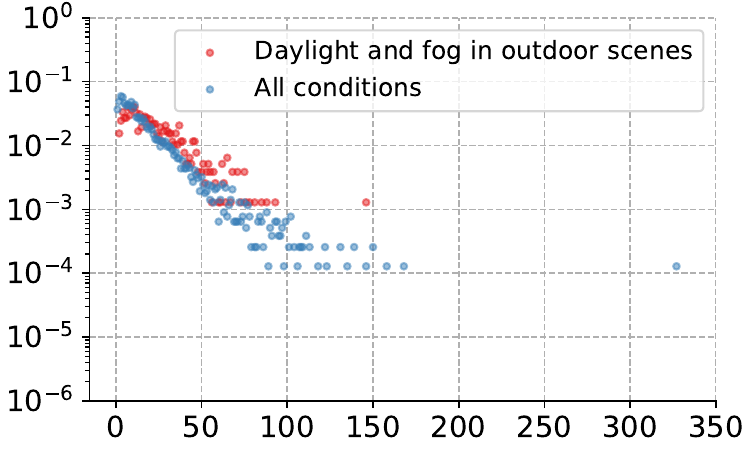}
      \put(-4, 9.5){\rotatebox{90}{\scriptsize Percent of images (\%)}}
    \end{overpic}
    \caption{Daylight and fog in outdoor scenes.}
    \label{fig:instance_per_image_fog}
  \end{subfigure}
  \begin{subfigure}[t]{0.26\textwidth}
    \begin{overpic}[width=\linewidth]{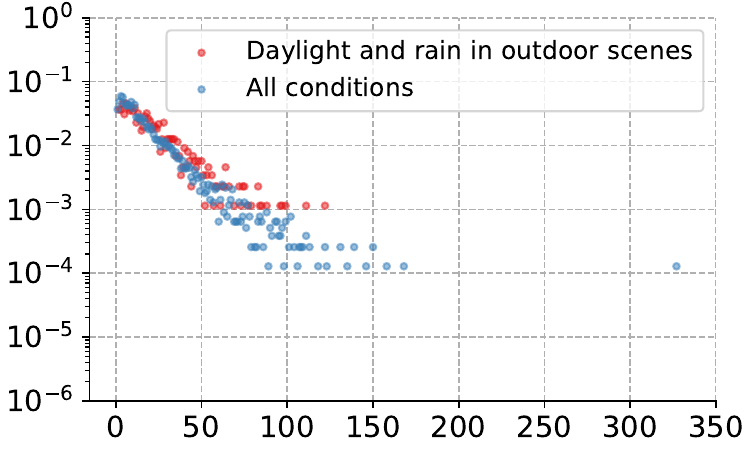}
      \put(-4, 9.5){\rotatebox{90}{\scriptsize Percent of images (\%)}}
    \end{overpic}
    \caption{Daylight and rain in outdoor scenes.}
    \label{fig:instance_per_image_rain}
  \end{subfigure}
  \begin{subfigure}[t]{0.26\textwidth}
    \begin{overpic}[width=\linewidth]{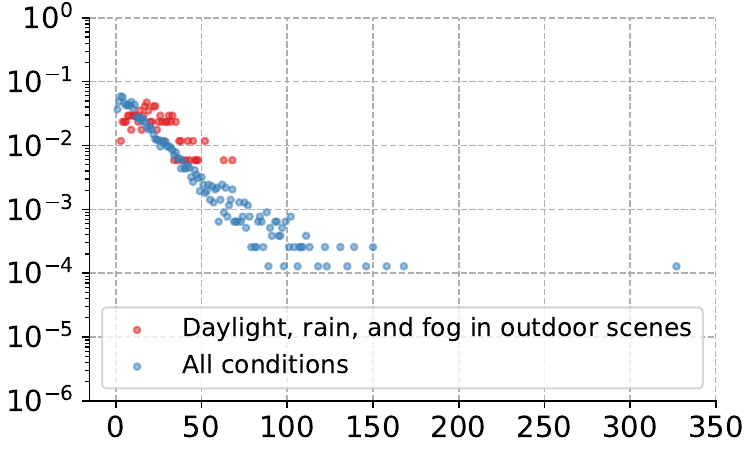}
      \put(-4, 9.5){\rotatebox{90}{\scriptsize Percent of images (\%)}}
    \end{overpic}
    \caption{Daylight, rain, and fog in outdoor scenes.}
    \label{fig:instance_per_image_rain_fog}
  \end{subfigure}
  \begin{subfigure}[t]{0.26\textwidth}
    \begin{overpic}[width=\linewidth]{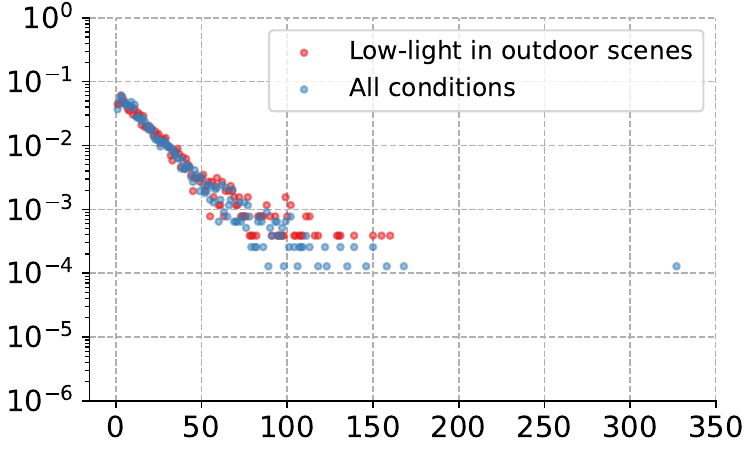}
      \put(-4, 9.5){\rotatebox{90}{\scriptsize Percent of images (\%)}}
    \end{overpic}
    \caption{Low-light in outdoor scenes.}
    \label{fig:instance_per_image_low_light}
  \end{subfigure}
  \begin{subfigure}[t]{0.26\textwidth}
    \begin{overpic}[width=\linewidth]{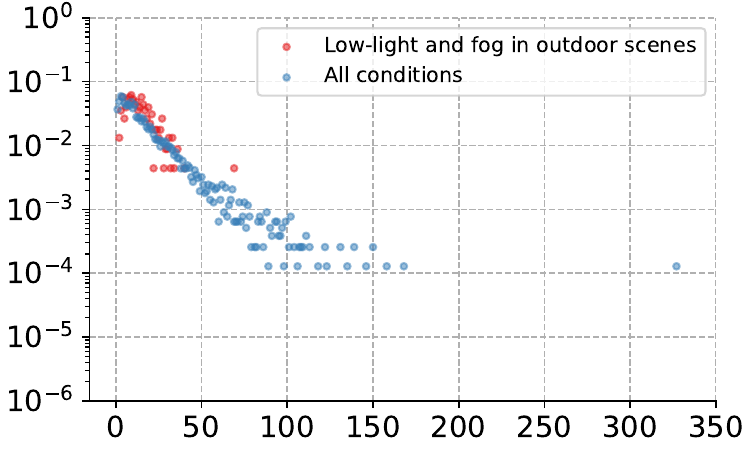}
      \put(-4, 9.5){\rotatebox{90}{\scriptsize Percent of images (\%)}}
    \end{overpic}
    \caption{Low-light and fog in outdoor scenes.}
    \label{fig:instance_per_image_low_light_fog}
  \end{subfigure}
  \begin{subfigure}[t]{0.26\textwidth}
    \begin{overpic}[width=\linewidth]{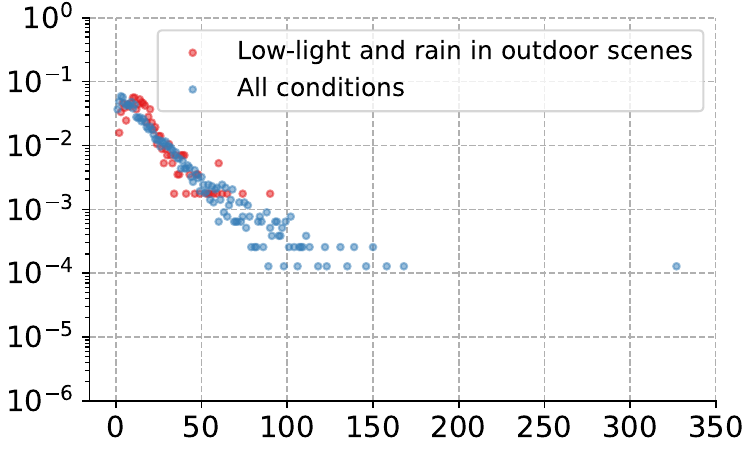}
      \put(-4, 9.5){\rotatebox{90}{\scriptsize Percent of images (\%)}}
    \end{overpic}
    \caption{Low-light and rain in outdoor scenes.}
    \label{fig:instance_per_image_low_light_rain}
  \end{subfigure}
  \caption{Distribution of the number of instances in images of each condition. 
  The horizontal axis represents the number of instances.
  }
  \label{fig:instance_per_image_per_condition}
\end{figure*}

\myPara{Instances per image.}
\figref{fig:instance_per_image_per_condition} shows the
distribution of the number of instances in images. 
In all conditions, there are complex images containing hundreds of instances.
The distribution of a single condition is close to the overall distribution, 
especially when the number of instances is less than 100. 
For cases exceeding 100, 
since there are fewer images in this range, 
there is some deviation between several conditions and the whole, 
\eg, the condition of low-light and fog in outdoor scenes, 
as shown in \figref{fig:instance_per_image_low_light_fog}. 

\begin{figure*}
  \centering
  \begin{subfigure}[t]{0.26\textwidth}
    \begin{overpic}[width=\linewidth]{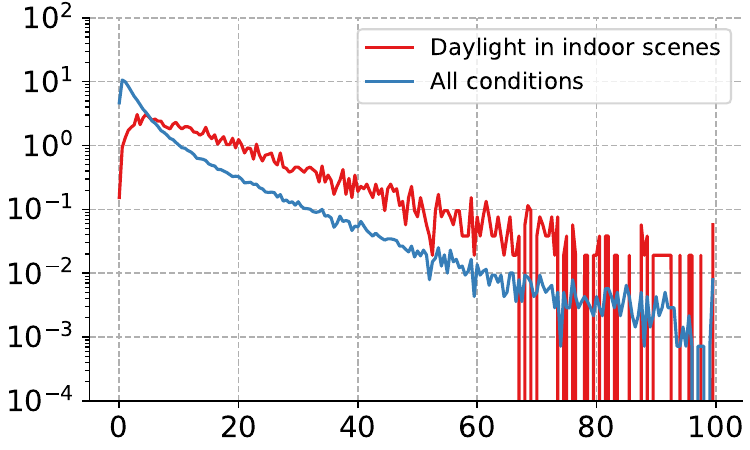}
      \put(-4, 7.0){\rotatebox{90}{\scriptsize Percent of instances (\%)}}
    \end{overpic}
    \caption{Daylight in indoor scenes.}
    \label{fig:instance_size_normal_light_indoor}
  \end{subfigure}
  \begin{subfigure}[t]{0.26\textwidth}
    \begin{overpic}[width=\linewidth]{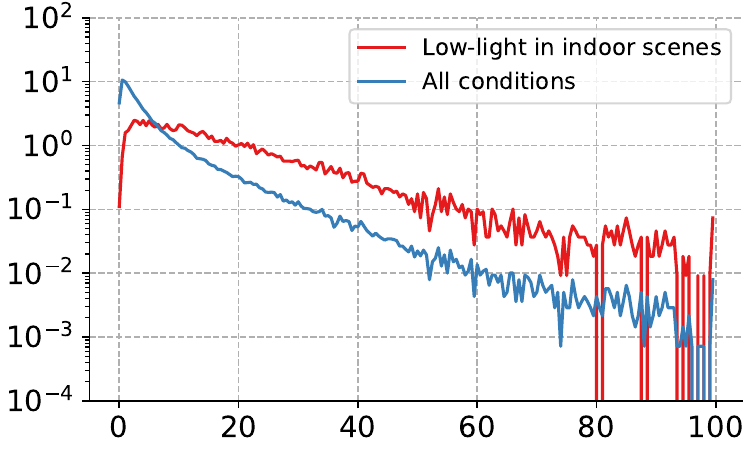}
      \put(-4, 7.0){\rotatebox{90}{\scriptsize Percent of instances (\%)}}
    \end{overpic}
    \caption{Low-light in indoor scenes.}
    \label{fig:instance_size_low_light_indoor}
  \end{subfigure}
  \begin{subfigure}[t]{0.26\textwidth}
    \begin{overpic}[width=\linewidth]{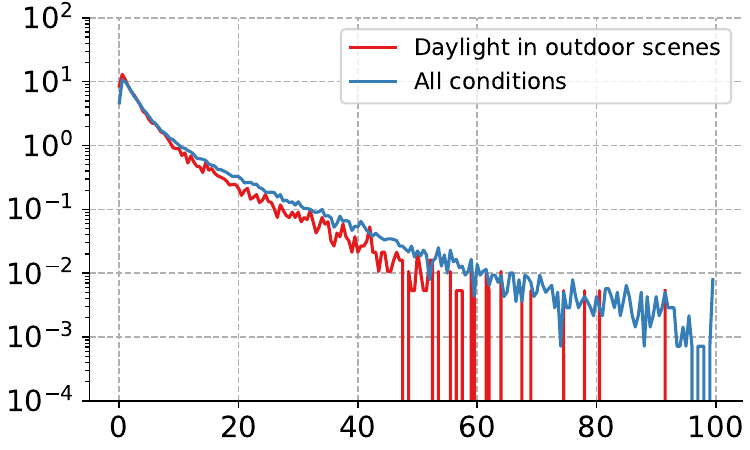}
      \put(-4, 7.0){\rotatebox{90}{\scriptsize Percent of instances (\%)}}
    \end{overpic}
    \caption{Daylight in outdoor scenes.}
    \label{fig:instance_size_normal_light}
  \end{subfigure}
  \begin{subfigure}[t]{0.26\textwidth}
    \begin{overpic}[width=\linewidth]{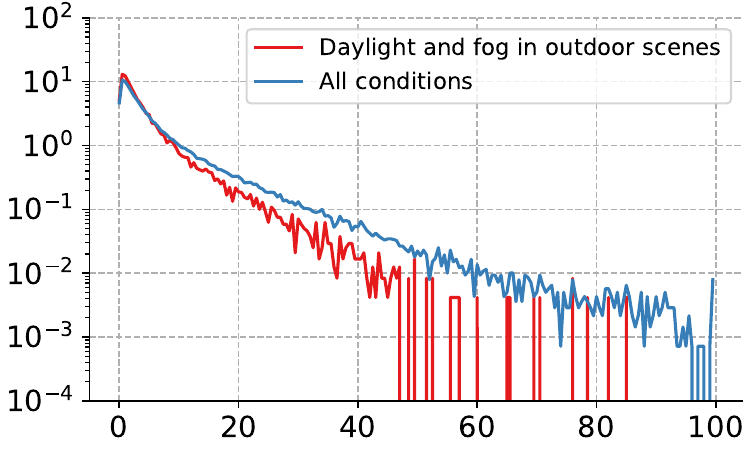}
      \put(-4, 7.0){\rotatebox{90}{\scriptsize Percent of instances (\%)}}
    \end{overpic}
    \caption{Daylight and fog in outdoor scenes.}
    \label{fig:instance_size_fog}
  \end{subfigure}
  \begin{subfigure}[t]{0.26\textwidth}
    \begin{overpic}[width=\linewidth]{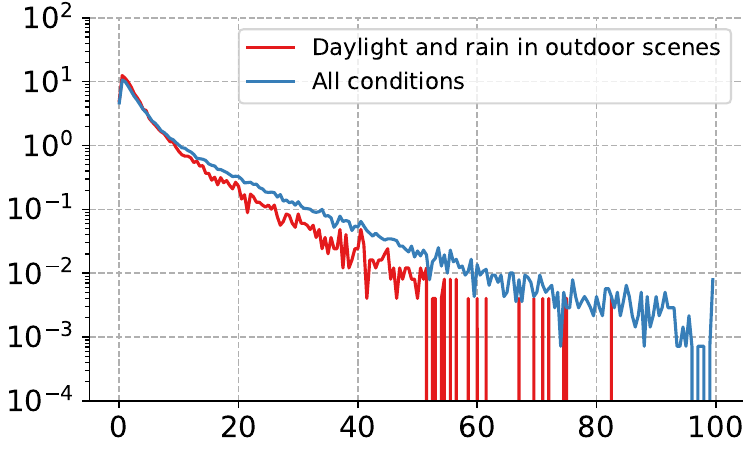}
      \put(-4, 7.0){\rotatebox{90}{\scriptsize Percent of instances (\%)}}
    \end{overpic}
    \caption{Daylight and rain in outdoor scenes.}
    \label{fig:instance_size_rain}
  \end{subfigure}
  \begin{subfigure}[t]{0.26\textwidth}
    \begin{overpic}[width=\linewidth]{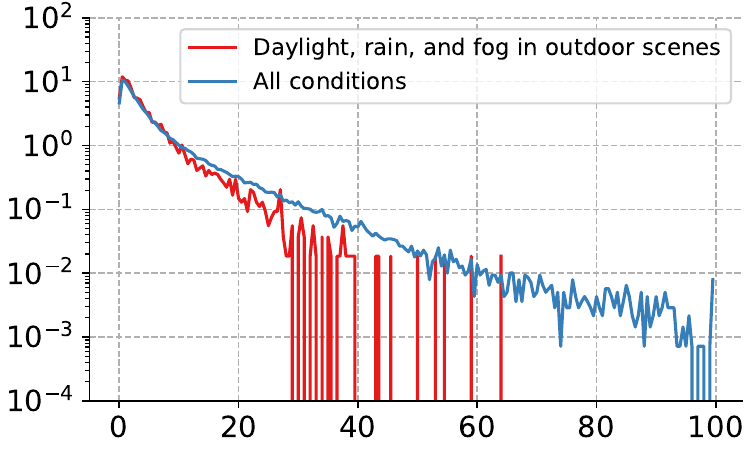}
      \put(-4, 7.0){\rotatebox{90}{\scriptsize Percent of instances (\%)}}
    \end{overpic}
    \caption{Daylight, rain, and fog in outdoor scenes.}
    \label{fig:instance_size_rain_fog}
  \end{subfigure}
  \begin{subfigure}[t]{0.26\textwidth}
    \begin{overpic}[width=\linewidth]{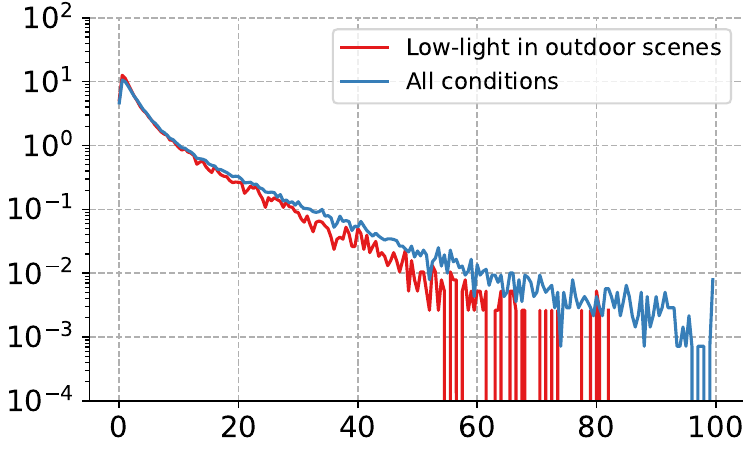}
      \put(-4, 7.0){\rotatebox{90}{\scriptsize Percent of instances (\%)}}
    \end{overpic}
    \caption{Low-light in outdoor scenes.}
    \label{fig:instance_size_low_light}
  \end{subfigure}
  \begin{subfigure}[t]{0.26\textwidth}
    \begin{overpic}[width=\linewidth]{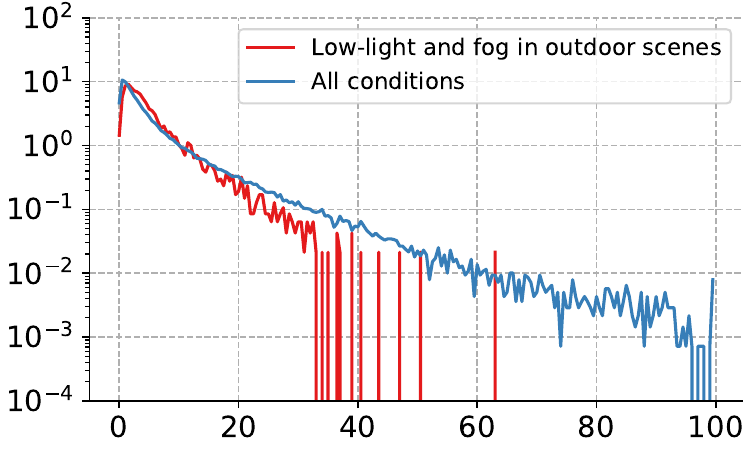}
      \put(-4, 7.0){\rotatebox{90}{\scriptsize Percent of instances (\%)}}
    \end{overpic}
    \caption{Low-light and fog in outdoor scenes.}
    \label{fig:instance_size_low_light_fog}
  \end{subfigure}
  \begin{subfigure}[t]{0.26\textwidth}
    \begin{overpic}[width=\linewidth]{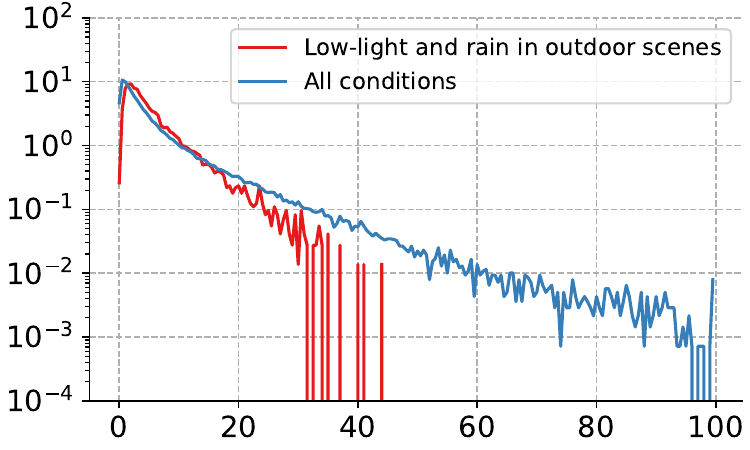}
      \put(-4, 7.0){\rotatebox{90}{\scriptsize Percent of instances (\%)}}
    \end{overpic}
    \caption{Low-light and rain in outdoor scenes.}
    \label{fig:instance_size_low_light_rain}
  \end{subfigure}
  \caption{Relative bounding box size  $\sqrt{\frac{\rm box\ area}{\rm image\ area}}$ of each condition. 
  The horizontal axis represents the relative bounding box size.
  }
  \label{fig:instance_size_per_condition}
\end{figure*}

\myPara{Bounding box size.}
\figref{fig:instance_size_per_condition} shows the distribution of the bounding box size for each condition, 
where small objects account for the majority in all conditions. 
Meanwhile, 
compared to the overall distribution of all conditions, 
the images in indoor scenes contain 
fewer small objects and more large objects, 
as shown in \figref{fig:instance_size_normal_light_indoor} and 
\figref{fig:instance_size_low_light_indoor}. 
In contrast, 
the outdoor scenes have a similar distribution to the overall distribution 
with a large number of small objects. 

\section{Experiments Settings}

Most hyperparameters follow the COCO dataset
in the mmdetection. 
For data augmentations, 
the images are resized between 800 and 1024 along the shorter side, 
while the longer side is no larger than 2048.
And the RandomFlip is used to augment images. 
For detectors using ConvNeXt, 
we set the layer-wise learning rate decay as 0.75 and 
the stochastic depth~(DropPath) ratio as 0.4. 
For detectors using Swin-T, 
the stochastic depth ratio is 0.2. 
For RAW pre-training, 
we follow the official codes released by ConvNeXt 
and Swin.

\begin{table}
  \centering
  \setlength{\tabcolsep}{5mm}
  \begin{tabular}{lccc}
    \toprule
    Distillation & AP & AP$_{\rm 50}$ & AP$_{\rm 75}$ \\
    \midrule
    \xmark & 34.1 & 52.4 & 35.9 \\
    Logit & 34.3 & 52.4 & 36.6 \\
    Logit + Feature & 34.8 & 53.3 & 36.7 \\
    \bottomrule
  \end{tabular}
  \caption{Ablation for the knowledge distillation when using Cascade RCNN and ConvNeXt-T.}
  \label{tab:distill}
\end{table}

\newcommand{\addexample}[1]{{\includegraphics[width=.28\textwidth]{figures/total_suppl_each_tag/#1}}}
\begin{table*}
  \centering
  \setlength{\tabcolsep}{1mm}
  \begin{tabular}{ccc}
    \toprule
    Indoor and Daylight & Indoor and Low-light & Outdoor, Daylight, and Clear Weather \\
    \midrule
    \addexample{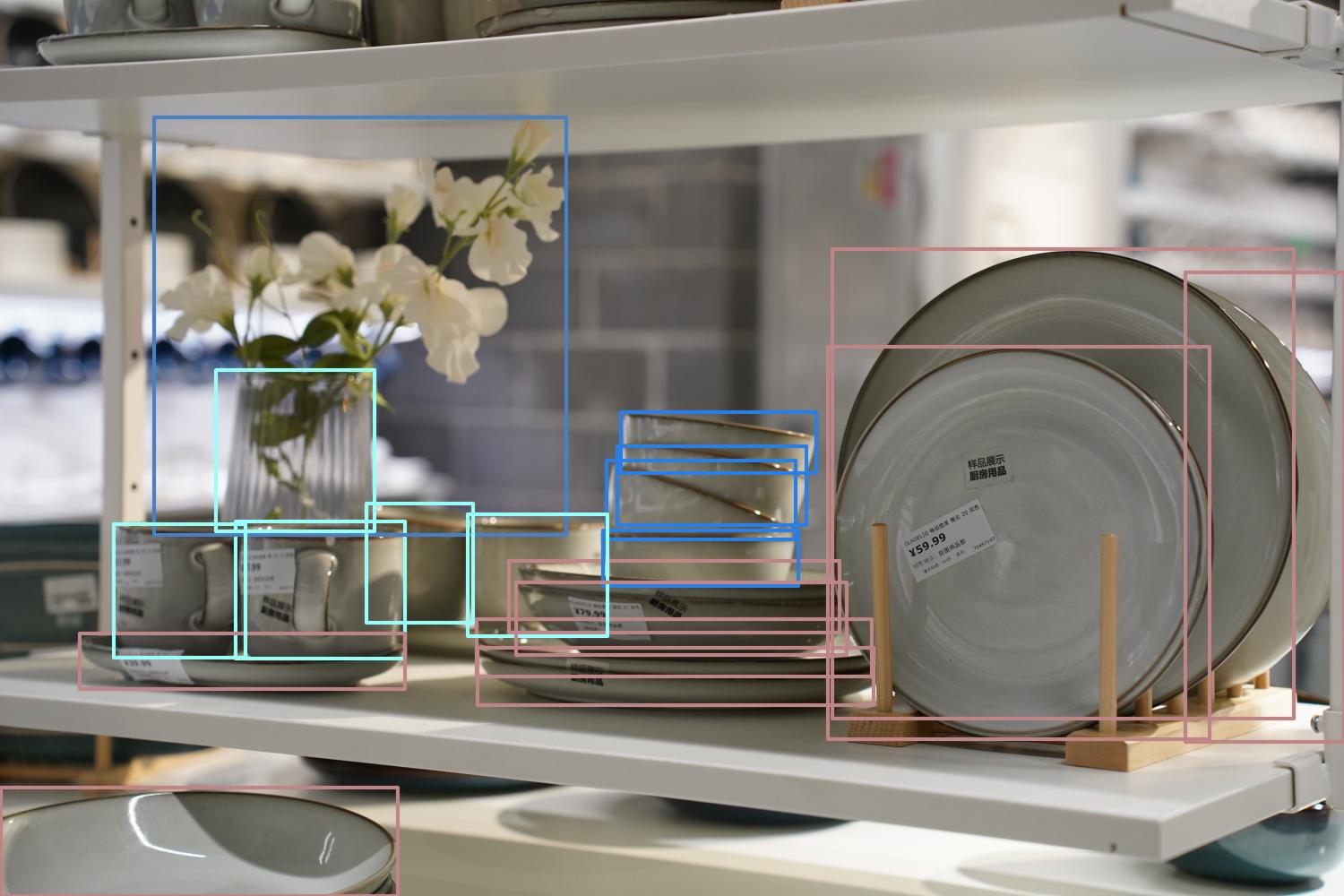} & \addexample{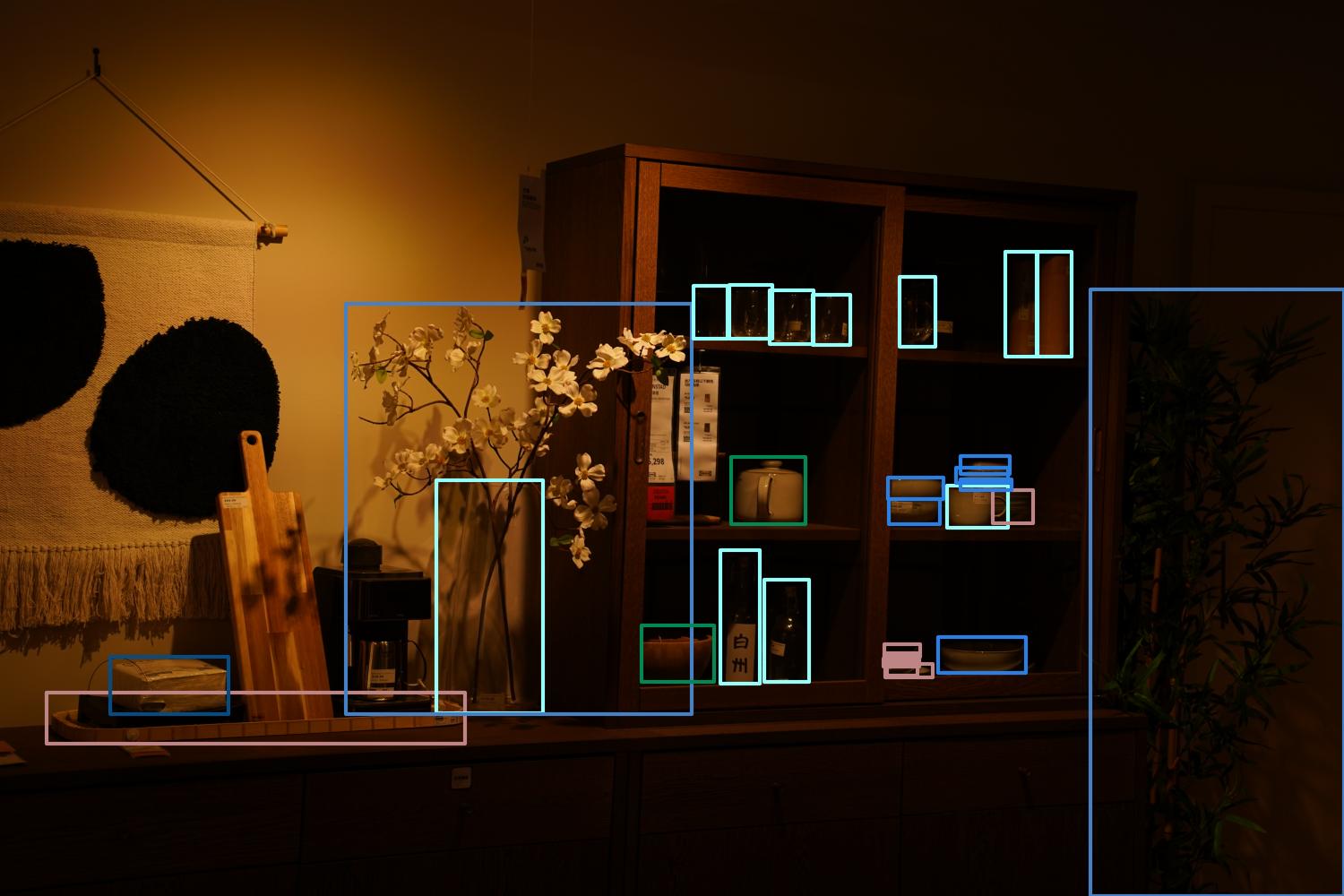} & \addexample{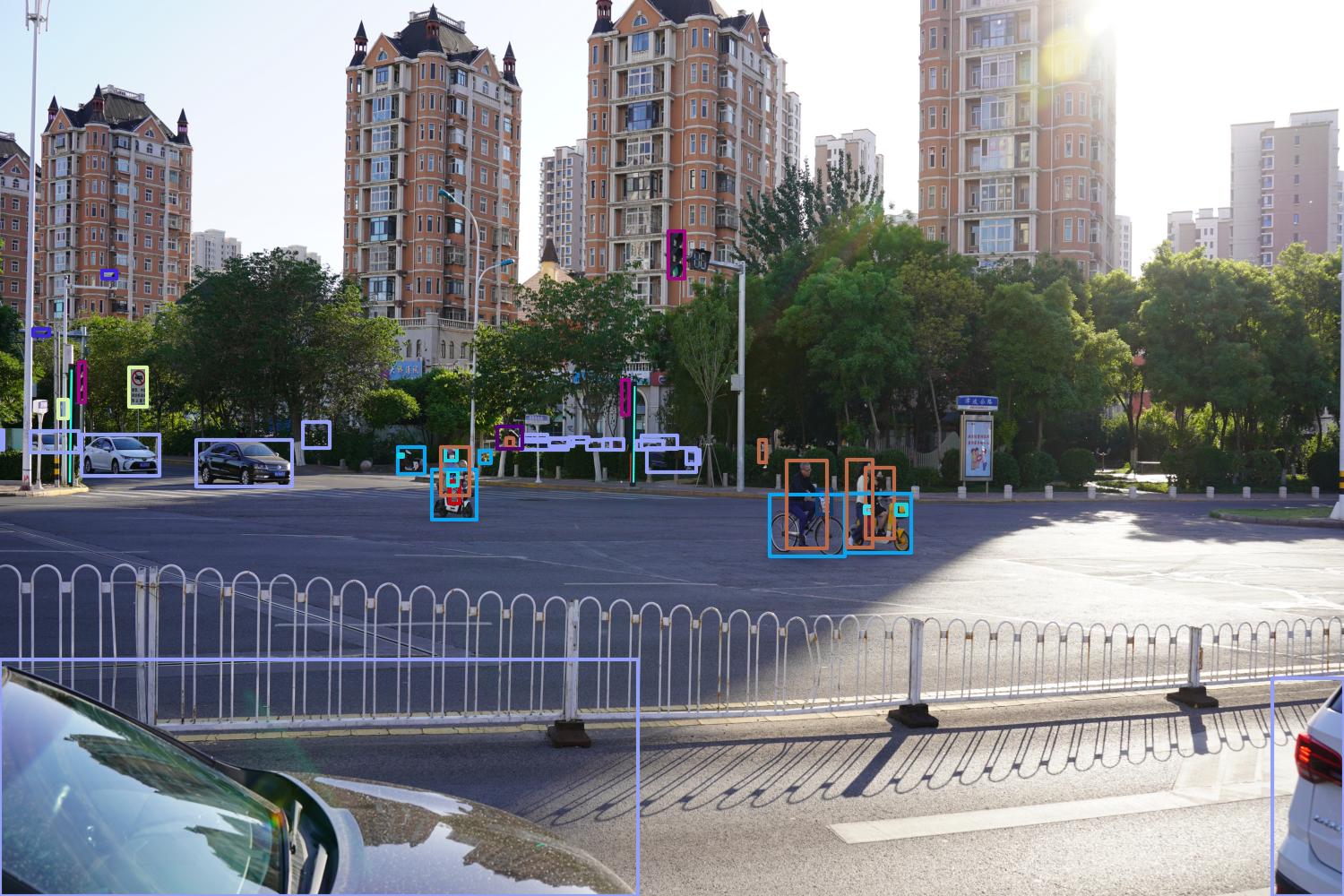} \\
    \midrule
    Outdoor, Daylight, and Fog Weather & Outdoor, Daylight, and Rain Weather & Outdoor, Daylight, and Rain+Fog Weather \\
    \midrule
    \addexample{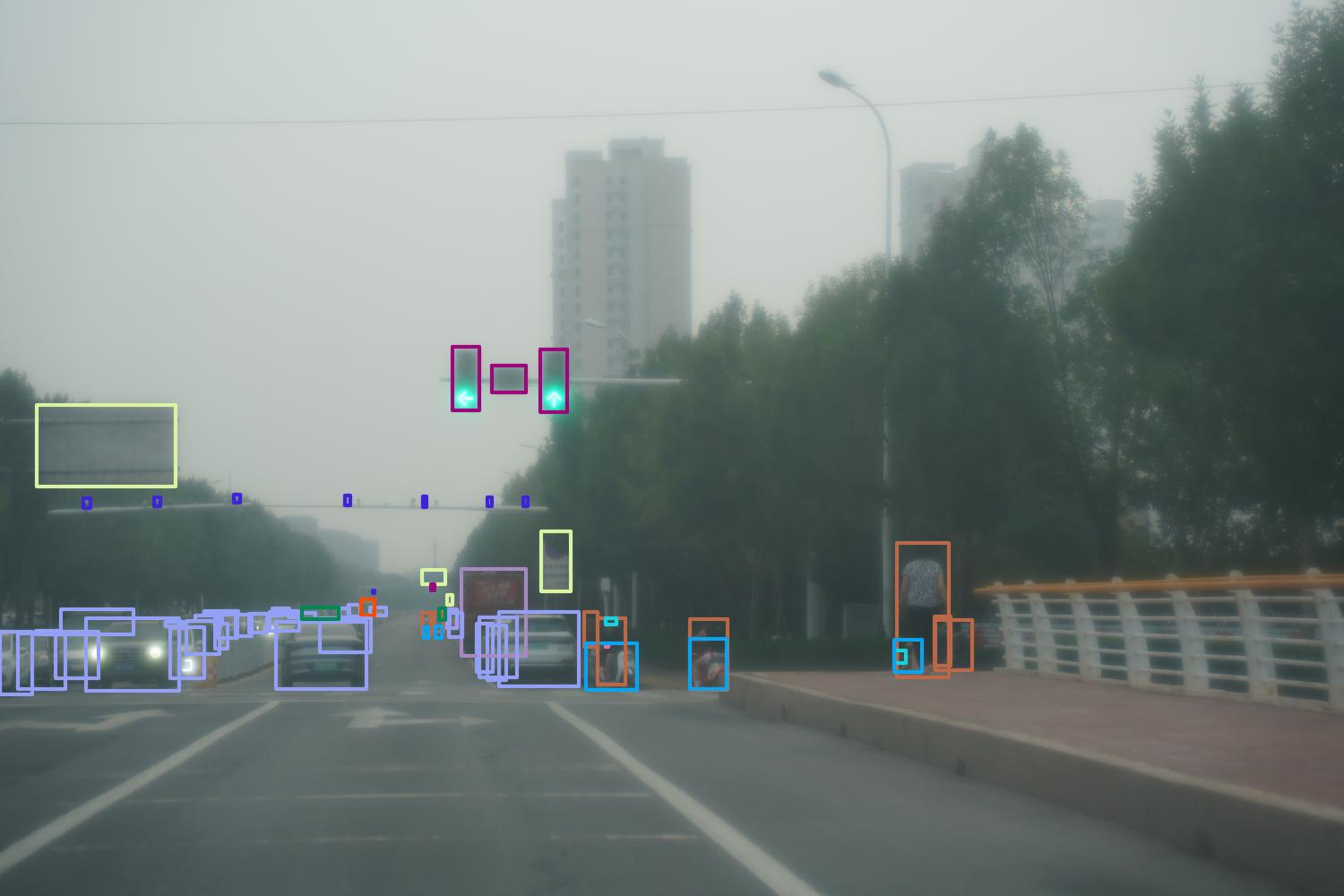} & \addexample{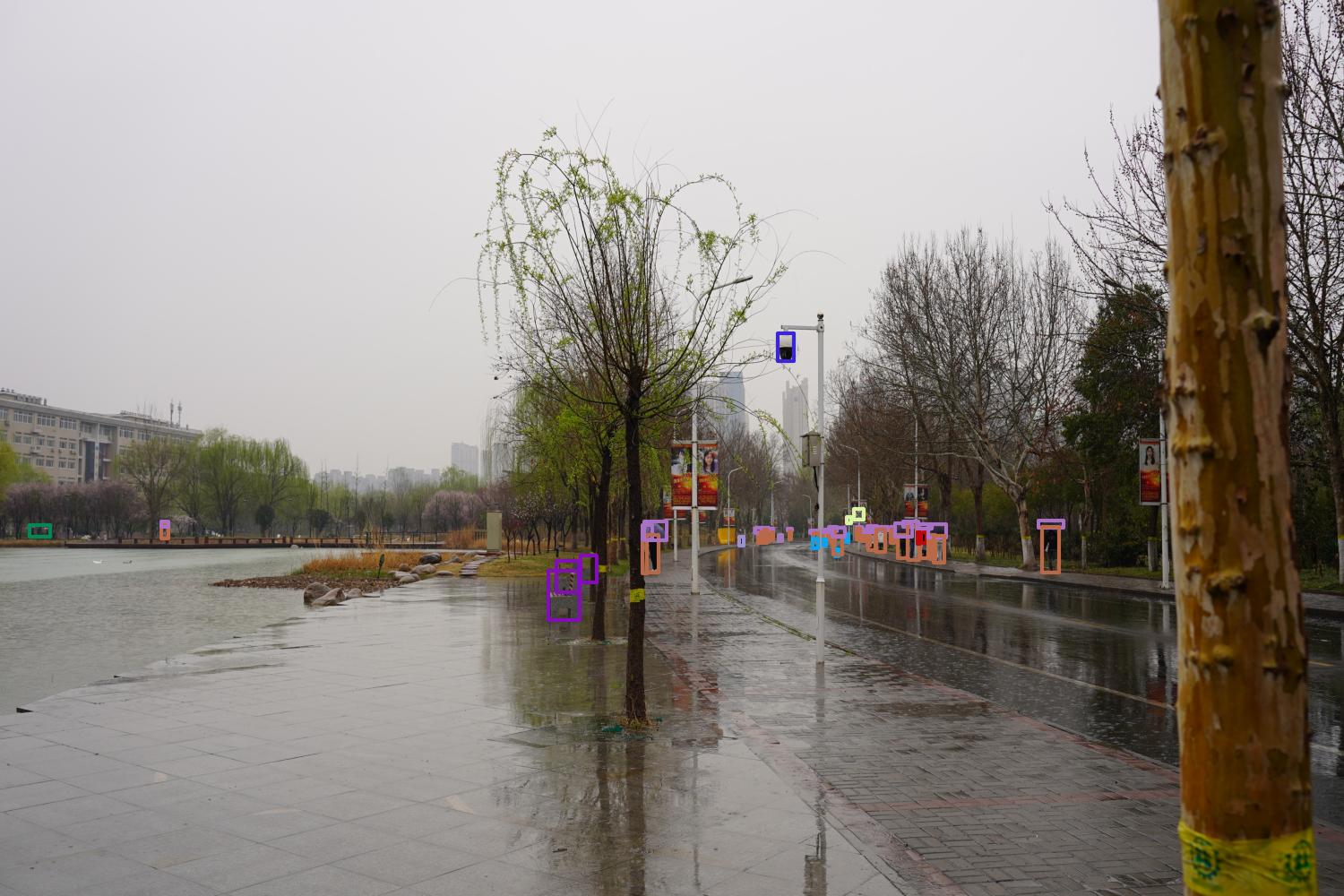} & \addexample{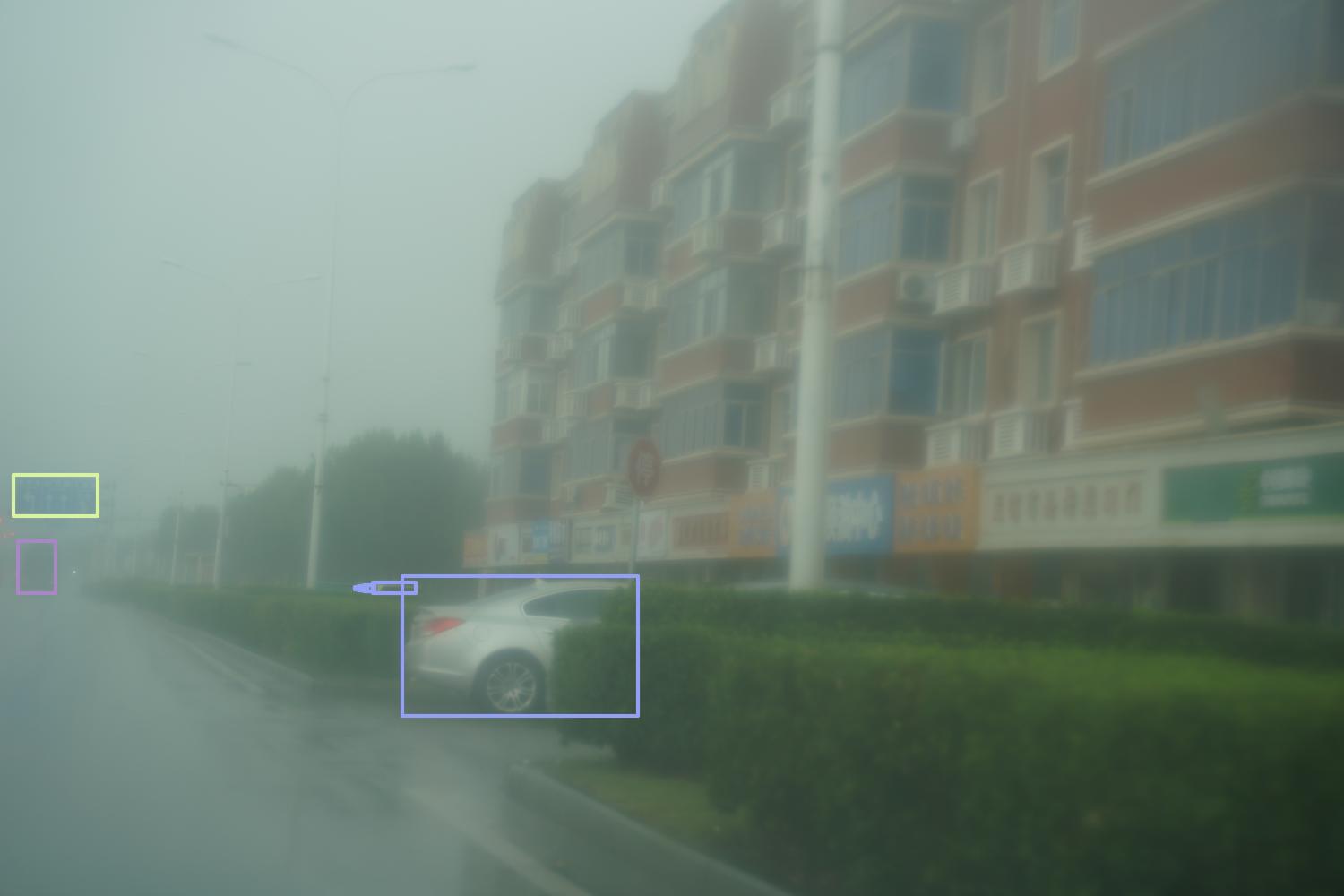}\\
    \midrule
    Outdoor, Low-light, and Clear Weather & Outdoor, Low-light, and Fog Weather & Outdoor, Low-light, and Rain Weather \\
    \midrule
    \addexample{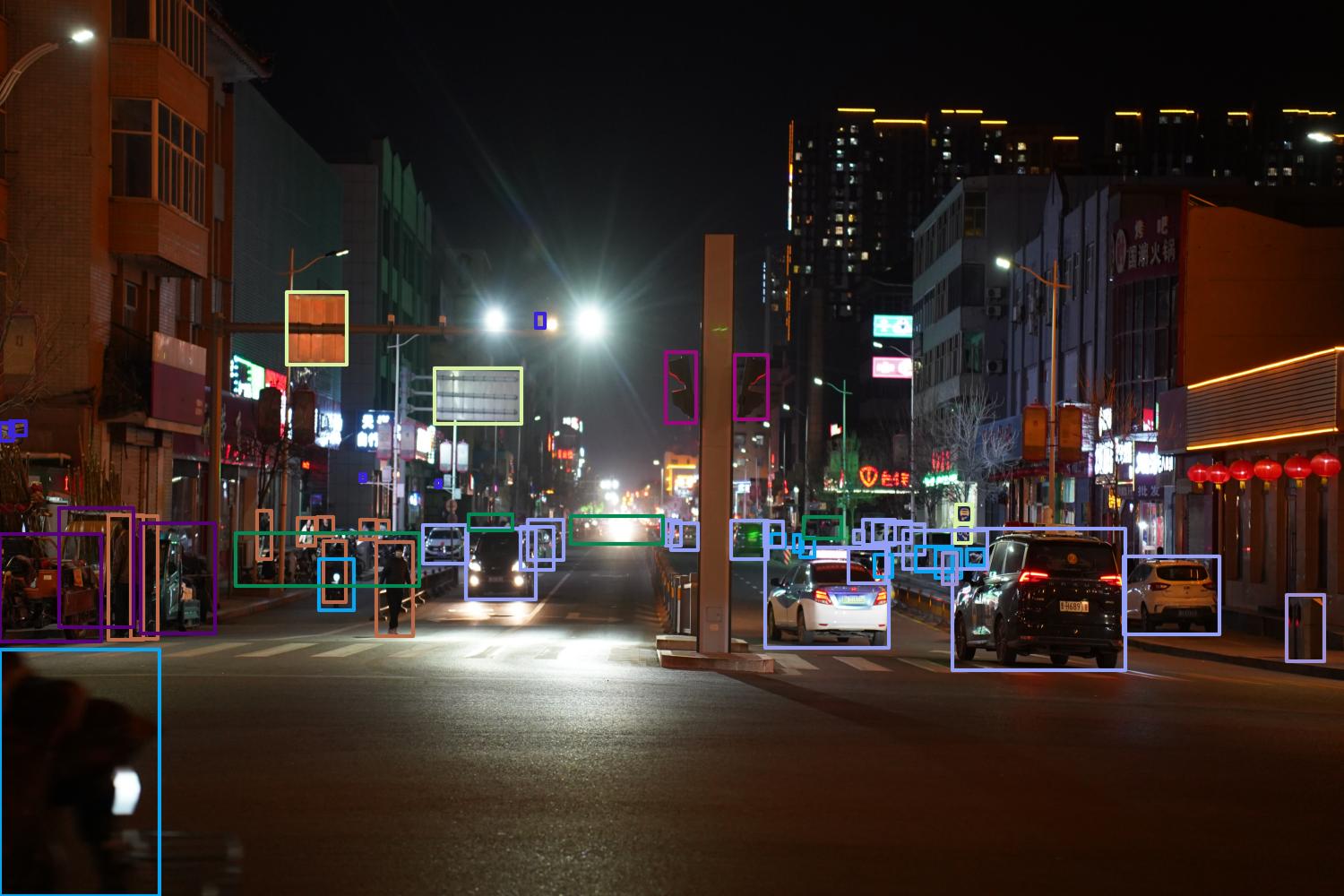} & \addexample{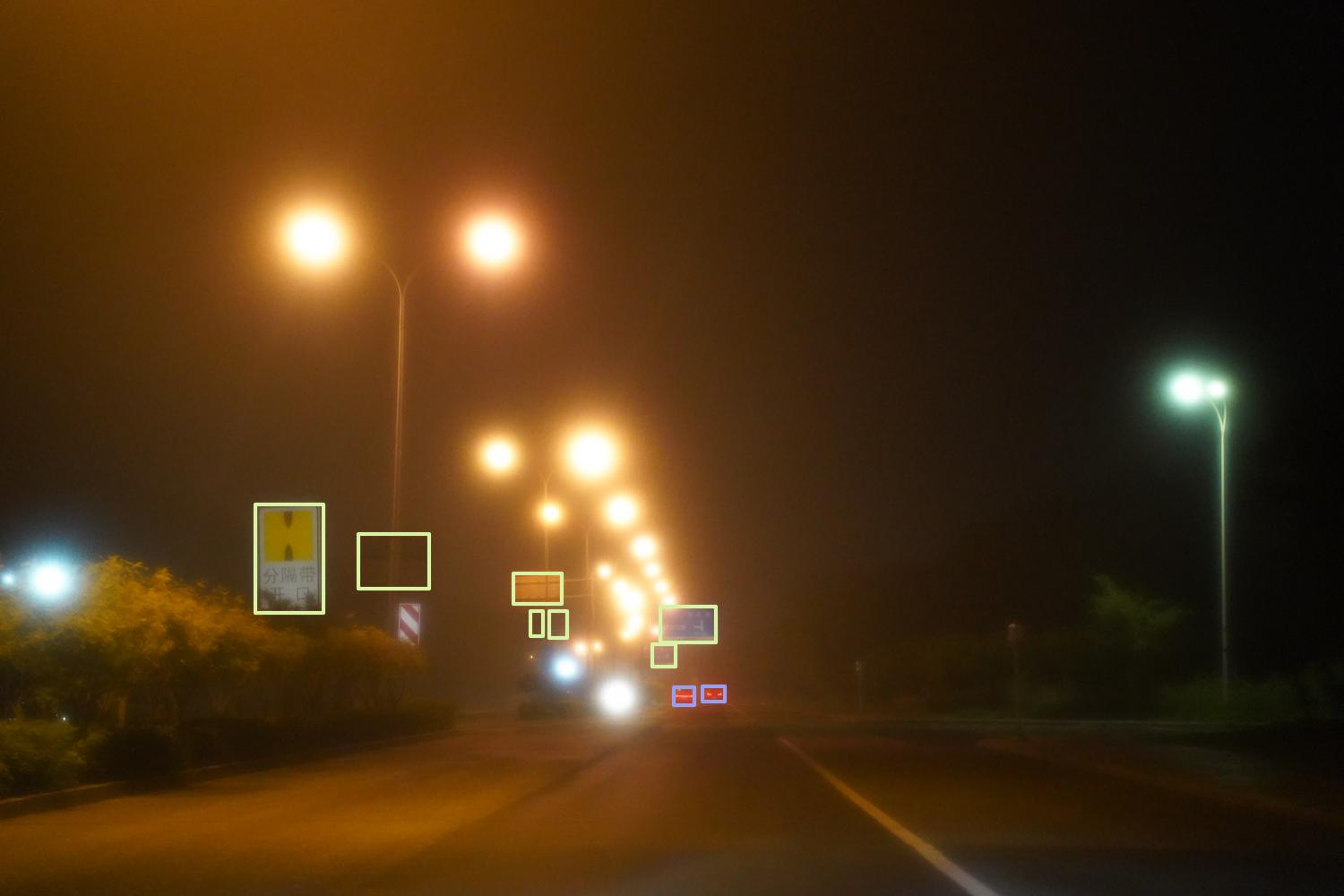} & \addexample{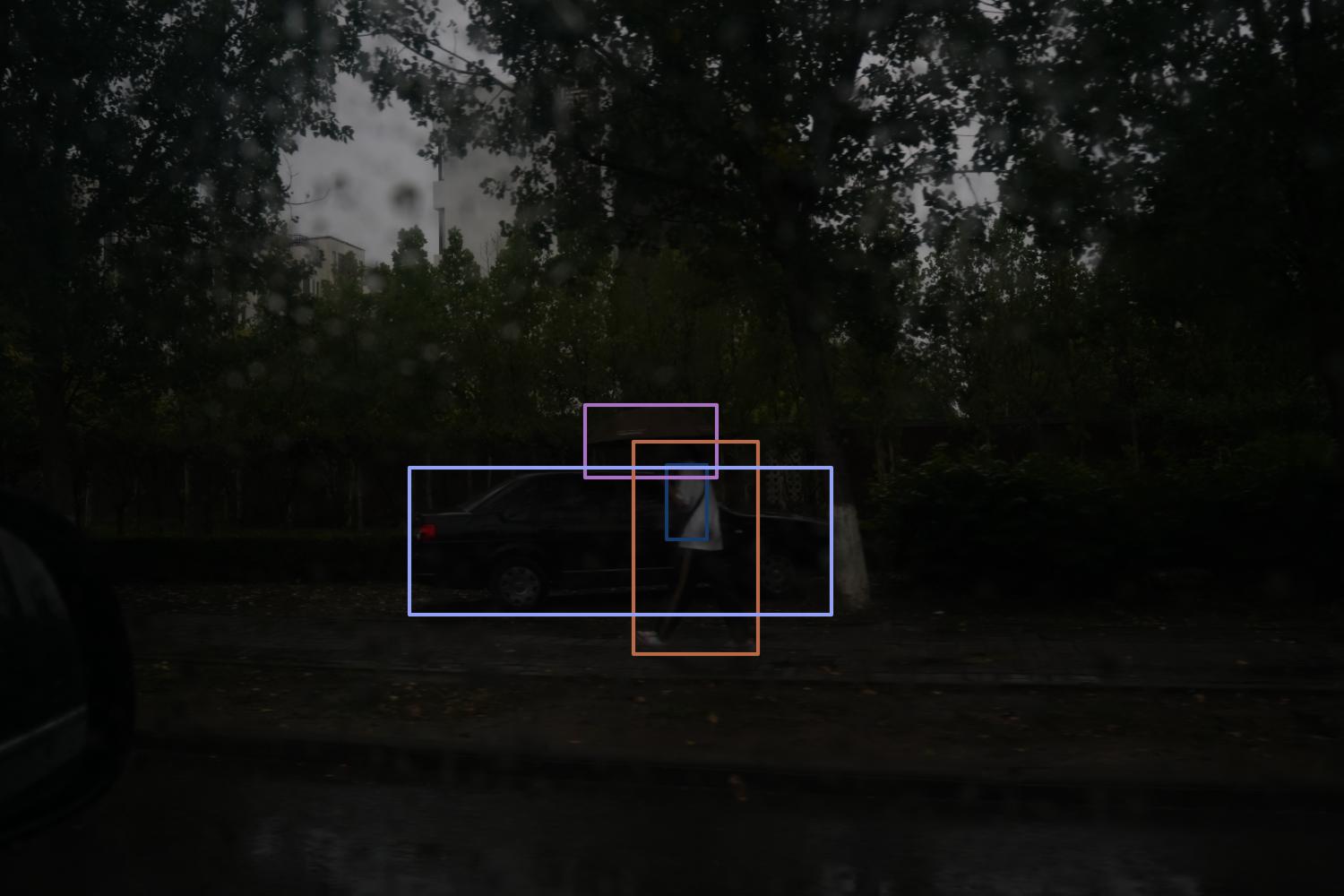} \\
    \bottomrule
  \end{tabular}
  \caption{The example of each condition in \nameofdata.}
  \label{tab:example_each_condition}
\end{table*}

\begin{table*}[t]
  \centering
  \setlength{\tabcolsep}{1.8mm}
  \begin{tabular}{lc|lc|lc|lc}
    \toprule
    Name & Instance & Name & Instance & Name & Instance & Name & Instance \\
    \midrule
    car & 42798 & person & 16864 & traffic sign & 13458 & surveillance camera & 9344 \\
    traffic light & 8884 & motorcycle & 7389 & truck & 4271 & chair & 3548 \\
    bottle/cup & 3168 & bicycle & 2134 & garbage can & 1993 & traffic cone & 1929 \\
    table & 1480 & tricycle & 1268 & helmet & 1211 & umbrella & 1142 \\
    pillow & 1055 & lamp & 1046 & handbag/satchel & 1001 & bus & 973 \\ 
    potted plant & 910 & hat & 893 & backpack & 818 & phone & 685 \\ 
    plate & 663 & vase & 639 & monitor & 586 & bus stop sign & 528 \\
    desk lamp & 431 & tent & 406 & sofa & 373 & clock & 333 \\ 
    bowl & 322 & pen & 228 & crane & 196 & wine glass & 191 \\
    bench & 191 & keyboard & 175 & mirror & 170 & bed & 158 \\
    mouse & 155 & pot & 151 & earphone & 151 & fire hydrant & 145 \\ 
    toilet paper & 138 & spoon & 105 & laptop & 104 & sink & 96 \\
    watch & 93 & fire extinguisher & 90 & suitcase & 85 & train & 72 \\
    dog & 64 & computer box & 58 & refrigerator & 43 & cans & 42 \\ 
    vending machine & 34 & airplane & 32 & boat & 29 & cat & 23 \\
    toilet & 20 & scissors & 19 \\
    \bottomrule
  \end{tabular}
  \caption{The annotated categories and the number of instances per category.
   }
  \label{tab:annotated_categories}
\end{table*}

\section{Distillation Implementation}

\myPara{Method.} 
Besides the supervised classification loss function, 
we use logit-based and feature-based distillation for cross-domain distillation. 
For logit-based distillation, 
we denote $z_s$ and $z_t$ as the output of student and teacher, respectively. 
$z_s$ and $z_t$ have been normalized by the SoftMax function. 
Then, the logit-based loss is calculated as follows:
\begin{equation}
  L_l = {\rm KLDivLoss}(y_s, y_t) = y_t\log{\frac{y_t}{y_s}}.
\end{equation}
For feature-based distillation, 
we denote $z_s$ and $z_t$ as the global feature output by the student and teacher, respectively. 
For ConvNeXt, 
the features are acquired 
through 
1) applying global average pooling to the output of the last block 
and 2) processing the feature using the last LayerNorm layer. 
The loss is calculated as follows:
\begin{equation}
  L_f = \frac{1}{C} \sum_{i=0}^{C-1}|z_s^i-z_t^i|, 
\end{equation}
where C means the number of dimensions.

\myPara{Ablation studies.}
We ablate the two types of distillation in \tabref{tab:distill}. 
Logit-based distillation improves the AP by 0.2\%, and 
feature-based distillation further extends the improvement by 0.5\%.

\end{document}